\PassOptionsToPackage{table,xcdraw}{xcolor}
\documentclass{article}

\PassOptionsToPackage{numbers}{natbib}

    \usepackage[preprint]{neurips_2025}

\usepackage[utf8]{inputenc} 
\usepackage[T1]{fontenc}    
\usepackage{hyperref}       
\usepackage{url}            
\usepackage{booktabs}       
\usepackage{amsfonts}       
\usepackage{nicefrac}       
\usepackage{microtype}      

\usepackage{microtype}
\usepackage{graphicx}
\usepackage{subcaption}
\usepackage{placeins}
\usepackage{amsmath}
\usepackage{amssymb}
\usepackage{mathtools}
\usepackage{amsthm}

\usepackage[textsize=tiny]{todonotes}

\usepackage[table,xcdraw]{xcolor}
\usepackage{pifont} 
\usepackage{multirow}
\usepackage{float}
\usepackage{float}

\usepackage{adjustbox}
\usepackage{array}    
\usepackage{siunitx}  
\newcommand{\best}[1]{\textbf{\textcolor{green!50!black}{#1}}}

\usepackage{wrapfig} 
\usepackage{algorithm}        
\usepackage{algpseudocode}    
\usepackage{hyperref}         
 
\newcommand{\cmark}{\ding{51}} 
\newcommand{\xmark}{\ding{55}} 
\newcommand{\method}{LwPTV }

\newtheorem{thm}{Theorem}
\newtheorem{proposition}{Proposition}

\newtheorem{rmk}{Remark}

\title{Merging Smarter, Generalizing Better: Enhancing Model Merging on OOD Data}

\author{%
  Bingjie Zhang\textsuperscript{1}, Hongkang Li\textsuperscript{2}, 
  Changlong Shi\textsuperscript{1}, Guowei Rong\textsuperscript{3},
  He Zhao\textsuperscript{4},
  Dongsheng Wang\textsuperscript{5}, \\
  \textbf{Dandan Guo\textsuperscript{1}} \thanks{Corresponding author.} \textbf{,}
  \textbf{Meng Wang\textsuperscript{2}}
  \\
  \textsuperscript{1}College of Artificial Intelligence, Jilin University \textsuperscript{2}Rensselaer Polytechnic Institute \\
  \textsuperscript{3}College of applied Science, 
Taiyuan University of Science and Technology\\
  \textsuperscript{4}CSIRO's Data61 
  \textsuperscript{5}College of Computer Science and Software Engineering, Shenzhen University\\
  \texttt{zhangbj24@mails.jlu.edu.cn}, 
  \texttt{lohek330@gmail.com}, \texttt{shicl22@mails.jlu.edu.cn}, \\ 
  \texttt{202118080119@stu.tyust.edu.cn}, \texttt{he.zhao@data61.csiro.au},
\texttt{wds\_dana@163.com}, \\ \texttt{guodandan@mails.jlu.edu.cn}, 
\texttt{wangm7@rpi.edu}
}

\begin{document}

\maketitle

\begin{abstract}
  Multi-task learning (MTL) concurrently trains a model on diverse task datasets to exploit common features, thereby improving overall performance across the tasks. Recent studies have dedicated efforts to merging multiple independent model parameters into a unified model for MTL, thus circumventing the need for training data and expanding the scope of applicable scenarios of MTL. However, current approaches to model merging predominantly concentrate on enhancing performance within in-domain (ID) datasets, often overlooking  their efficacy on out-of-domain (OOD) datasets. In this work, we proposed \method (Layer-wise Pruning Task Vector) by building a saliency score, measuring the redundancy of parameters in task vectors. Designed in this way ours can achieve mask vector for each task and thus perform layer-wise pruning on the task vectors, only keeping the pre-trained model parameters at the corresponding layer in merged model. Owing to its flexibility, our method can be seamlessly integrated with most of existing model merging methods to  improve their performance on OOD tasks. Extensive experiments demonstrate that the application of our method results in substantial enhancements in OOD performance while preserving the ability on ID tasks.
\end{abstract}

\vspace{-3mm}
\section{Introduction}
\label{sec:Introduction}

Pre-trained models (PTMs) constitute a cornerstone of deep learning, supporting numerous contemporary methodologies by virtue of their capacity to extract generalized features from vast data repositories \cite{bommasani2021opportunities, zhuang2020comprehensive}. Typically, the fine-tuning of pre-trained models with task-specific data is employed to enhance performance \cite{shnarch2022label, kenton2019bert}. This process yields numerous checkpoints originating from the PTMs \cite{Poth_Pfeiffer_Rücklé_Gurevych_2021}.  Fine-tuning individual models for each task incurs substantial storage and deployment costs \cite{dettmersqlora}. Multi-task learning (MTL) offers an alternative by training a single model on multiple tasks, reducing storage demands. However, MTL introduces significant computational overhead and is constrained by privacy concerns due to the necessity of aggregating diverse datasets \cite{jindataless}.
Recently, the trend in research has shifted to an alternative paradigm: merging various individual fine-tuned models into one single multi-task model, bypassing the requirement for the original training data. This approach, known as model merging, addresses these limitations by merging model parameters rather than engaging in further training \cite{wortsman2022model, matena2022merging, jindataless}.

\begin{wrapfigure}[15]{R}{7cm}
\vspace{-0.8em}
    \centering
    \vspace{-2mm}
    \includegraphics[scale=0.23]{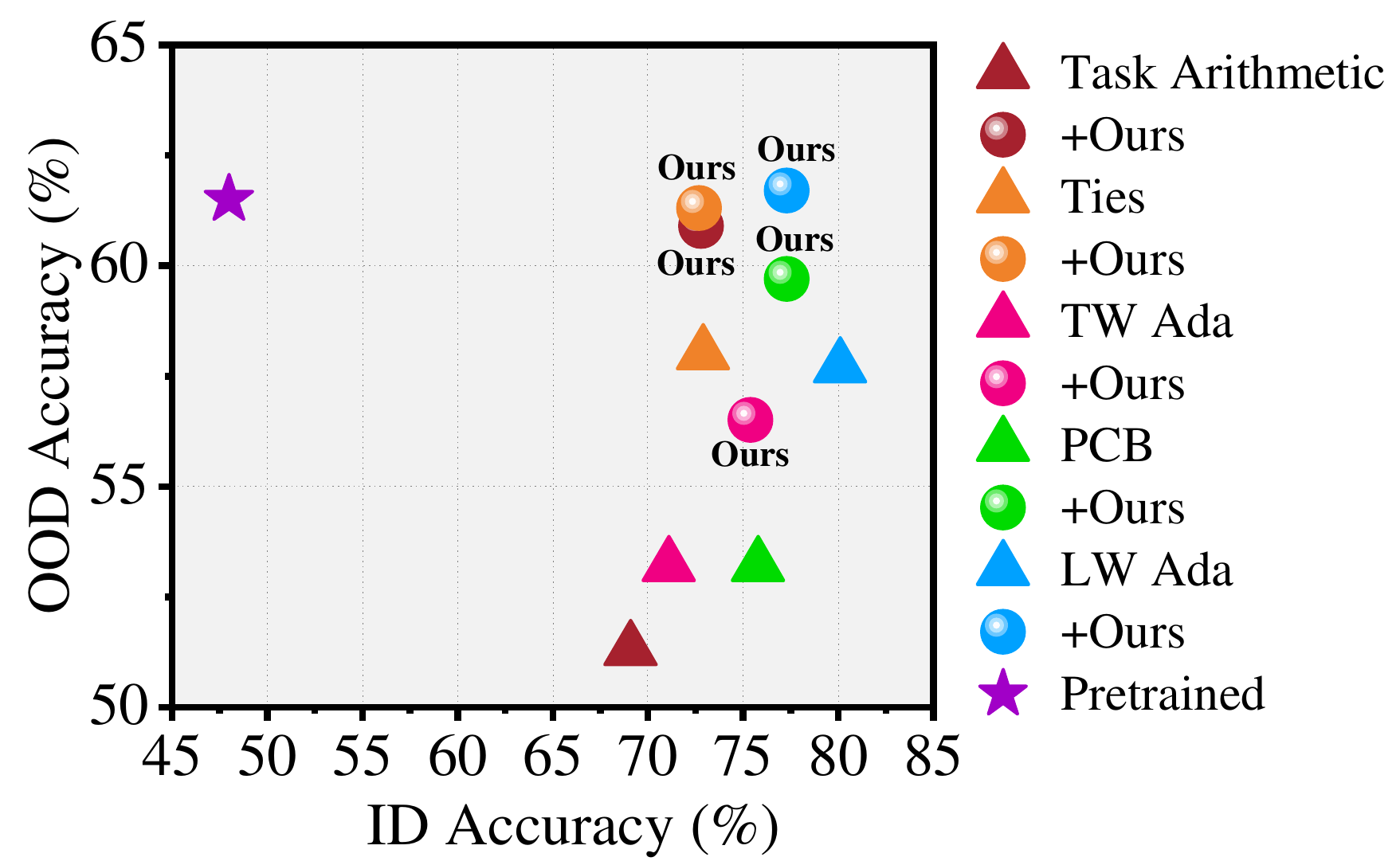}
    \captionsetup{font={small}}
    \vspace{-2mm}
    \caption{\small{ID and OOD performance of model merging methods on ViT-B/32; see ID and OOD tasks  in Experiments.
   Triangle and circle with same color mean baseline and baseline+ ours.
    }}
    \label{fig:ID_OOD}
\end{wrapfigure}


The most typical technique is Weighted Averaging, which directly averages the weights of multiple fine-tuned models \cite{wortsman2022model}. However, averaging often leads to a significant degradation in performance. To this end, various techniques, including Fisher Merging \cite{matena2022merging}, RegMean\cite{jindataless}, and Task Arithmetic \cite{ilharco2022editing}, have been proposed. Although these techniques improve performance, they primarily focus  on improving accuracy within ID datasets - the datasets used to fine-tune task-specific models. However, real-world applications often encounter data distributions that differ from those seen during training. The ability of a model to perform well in out-of-domain (OOD) data directly impacts its reliability, robustness, and deployment in dynamic environments such as healthcare, autonomous systems. Despite its importance, OOD robustness remains an underexplored aspect in model merging.
As shown in Fig.\ref{fig:ID_OOD}, although having better performance on in-domain (ID) data, merged models with current merging methods usually exhibit significant performance degradation on OOD tasks when compared with pre-trained models, which have stronger generalization capabilities.  The primary reason for the performance drop  lies in: fine-tuning updates the parameters of pre-trained models to fit ID datasets, often at the cost of perturbing generalizable features \cite{kumarfine, zhumodel}. 
Therefore, a better balance is needed between preserving generalization and incorporating task-specific adaptations.

In this work, we investigate the feasibility of pruning parameters of task vectors and only keeping the corresponding parameters in pre-trained models to enhance the generalization performance on OOD data of merged models. The central challenge lies in designing a principled criterion for identifying and removing task vector components that do not contribute meaningfully to {ID tasks}. Recall that task vectors can be interpreted as perturbations that align pre-trained models to specific tasks~\cite{yadav2023resolving,yangadamerging, yangrepresentation, ilharco2022editing, du2024parameter}. Shown in \cite{LZZC25}, these vectors encode both discriminative and redundant information. With in-depth analysis, we posit that task-specific discriminative features tend to be diverse across tasks, while redundant or low-signal features exhibit consistency across task vectors, indicating low task-specific relevance. This motivates the hypothesis that shared patterns in task vectors reflect low-salience, potentially non-discriminative modifications that do not meaningfully contribute to performance, especially in OOD settings. More interestingly, we find that this diversity of task-specific discriminative features is not uniformly distributed across the model, but rather exhibits a layer-wise structure. 
Based on these insights, we propose an adaptive method, \method (\textbf{L}ayer-\textbf{w}ise \textbf{P}runing \textbf{T}ask \textbf{V}ector for model merging). 
We define a layer-wise saliency score, quantifying the deviation of a given task vector from the layer-wise mean across tasks. Formally, for each parameter in a layer, this score captures the absolute difference between its task-specific value and the mean across task vectors. A low saliency score implies the parameter shift is consistent across tasks and thus likely redundant, motivating its removal in favor of the original pre-trained value. We construct layer-wise masks based on this score and additionally define a shared mask to preserve ID performance by retaining critical task-specific adaptations across all tasks.
Ours can be used a plug-and-play approach to enhance OOD generalization of most of model merging methods due to its flexibility. Extensive experiments prove that ours effectively maintains the pre-trained model’s generalization ability while leveraging task-specific adaptations, leading to improved OOD robustness while preserving ID performance.




\vspace{-2mm}

\section{Related work}
\label{sec:RelatedWork}
\vspace{-2mm}
\textbf{Multi-task Learning.} MTL involves training a model on data from multiple tasks simultaneously, with the aim of leveraging shared representations to enhance performance across all tasks \cite{vandenhende2021multi, zhang2023uni3d}. However, when confronted a new task, MTL requires access to labeled data from multiple tasks to train from scratch. This leads to high training costs and limited data accessibility due to privacy concerns \cite{pruksachatkun2020intermediate, poth2021pre, weller2022use, fifty2021efficiently}. 


\textbf{Model Merging.} Model merging \cite{wortsman2022model, matena2022merging, ilharco2022editing, yadav2023resolving} aims to combine several fine-tuned task-specific models into a single, unified multi-task model, by utilizing the existing task-specific model weights without requiring additional training. Model merging can be classified into two types. The first type involves merging multiple models that were trained on the same task, with the goal of improving the overall generalization of the model \cite{guptastochastic, cha2021swad, ainsworthgit, singh2020model}. Another type focuses on merging models for different tasks in order to enable MTL \cite{wortsman2022model, ilharco2022editing, yadav2023resolving, yangrepresentation}. 

The second type is our focus. A naive method is model averaging technique, which often leads to significant performance degradation \cite{wortsman2022model}. To address the performance gap between fine-tuned models and the merging model in ID tasks, various methodologies have been proposed. 
Fisher Merging \cite{matena2022merging} and RegMean \cite{jindataless} enhance model merging by utilizing Fisher information matrices and inner-product matrices, respectively, to calculate weighted coefficients for individual models. 
Task Arithmetic \cite{ilharco2022editing} merges models introduce task vector, defined as the parameter differences between fine-tuned models and the pre-trained model, which is effective in model merging. As a follow up, some works are further proposed. For example, Ties-Merging \cite{yadav2023resolving} aims to tackle the task conflicts among multiple models; AdaMerging \cite{yangadamerging} introduces the adaptive learning about the merging coefficients; Surgery method \cite{yangrepresentation} addresses representation bias  between merged model and task-specific models; DARE \cite{yu2024language} introduces a preprocessing step called drop and rescale, which reduces interference by randomly eliminating most elements and rescaling the remaining ones in each task vector before merging fine-tuned LLMs; WEMoE \cite{tangmerging} dynamically combines shared and task-specific knowledge based on the input sample; PCB-MERGING \cite{du2024parameter} effectively addresses parameter competition by  adjusting parameter coefficients during the merging process.
 
 The methods discussed above focus on enhancing the ID performance of merging model. While Ties-Merging, AdaMerging, WEMoE, and PCB evaluate their performance on OOD data, their primary goal is still the ID tasks.
 In contrast, we aim to improve merging model on OOD tasks while preserving the ability to tackle the ID tasks. Besides, ours can be achieved only based on task vectors and can be combined with  existing task vector-based methods in a flexible way.

\textbf{Out-of-Distribution.} A challenge commonly referred to as OOD generalization, continues to pose a substantial challenge in the field of machine learning. Despite the remarkable zero-shot capabilities demonstrated by large pre-trained models, such as CLIP \cite{li2022clip, radford2021learning}, further finetuning on downstream tasks could potentially lead to decreased performance with OOD data \cite{nguyen2024saft, kumarfine, chen2023fine, shuttleworth2024lora}. Recent studies have proposed methods to mitigate this issue. 
 WiSE-FT \cite{wortsman2022robust} enhances the OOD performance of fine-tuned models by performing linear interpolation between the fine-tuned model and its corresponding pre-trained model.  Model Stock exploits the anchoring effect of pretrained models and the geometric properties of fine-tuning parameters to utilize two fine-tuned models, obtained through optimization with different random seeds for the same task, to approximate the center of the parameter distribution \cite{jang2024model}. 
LiNeS \cite{wang2024lines} employs a depth-dependent scaling strategy for parameter updates. These strategys aims to enhance the fine-tuned model’s performance on  OOD tasks. Different from them, ours aims to enhance the generalization capabilities of merging models, where we design the plug-and-play saliency score to prune redundant parameters of task vectors in a layer-wise manner.

\vspace{-2mm}
\section{Preliminaries}
\label{sec:ProblemFormulation}

\vspace{-2mm}


\textbf{Problem setup.} Denote the pre-trained model as $f(\mathbf{x};\boldsymbol{\theta}_{pre})$, where $\boldsymbol{\theta}_{pre} = \{\boldsymbol{\theta}_{pre}^1, ...\boldsymbol{\theta}_{pre}^l,...\boldsymbol{\theta}_{pre}^L\}$, $L$ is the number of layers and 
$\boldsymbol{\theta}_{pre}^l$ is the parameter of the $l$-th  layer.
We aim to fine-tune the model on $K$ downstream tasks $\{T_k\}_{k=1}^K$ to get $K$ finetuned models $\{f(\mathbf{x};\boldsymbol{\theta}_{K})\}_{k=1}^{K}$, where each finetuned model has the same size parameters with pretrained model and $\boldsymbol{\theta}_{k} = \{\boldsymbol{\theta}_{k}^1, ...\boldsymbol{\theta}_{k}^l,...\boldsymbol{\theta}_{k}^L\}$.
Model merging aims to combine the weights $\{\boldsymbol{\theta}_k\}^K_{k=1}$ into a new set of weights $\boldsymbol{\theta}_m$, enabling $f(\mathbf{x};\boldsymbol{\theta}_m)$ to perform $K$ tasks  without retraining using task-specific data. 

\textbf{Task vector-based methods.} A recent study \cite{ilharco2022editing} proposed the idea of ``task vectors'' for model merging, which has been further explored by subsequent research \cite{yadav2023resolving,yangadamerging}. A task vector $\boldsymbol{\tau}_k$ is defined as the difference between the fine-tuned model parameters $\boldsymbol{\theta}_k$ and the initial pre-trained parameters $\boldsymbol{\theta}_{pre}$. Specifically, for  the $k$-th task, its task vector is given as:
\begin{eqnarray}
\boldsymbol{\tau}_k=\boldsymbol{\theta}_k-\boldsymbol{\theta}_{pre}, \quad \boldsymbol{\tau}_k =\{\boldsymbol{\tau}_{k}^1, ...\boldsymbol{\tau}_{k}^l,...\boldsymbol{\tau}_{k}^L\}.
\label{Eq3-1}
\end{eqnarray}

Now, based on the task vectors, the merged function can be expressed as $\mathcal{F(\cdot)}$ as follows:
\begin{eqnarray}
 {\boldsymbol{\theta}}_{m}=\mathcal{F}(\boldsymbol{\theta}_{pre},\boldsymbol{\tau}_{1},\boldsymbol{\tau}_{2},\cdots, \boldsymbol{\tau}_{K}).
\label{Eq3-2}
\end{eqnarray}
For example, in terms of Task Arithmetic, the merged model weights can be expressed as $\boldsymbol{\theta}_m=\boldsymbol{\theta}_{pre}+\lambda\sum_{k=1}^K  \boldsymbol{\tau}_k$, where $\lambda$ denotes the merging coefficient. As for Task-wise AdaMerging, the merged model weights are computed as $\boldsymbol{\theta}_m=\boldsymbol{\theta}_{pre}+\sum_{k=1}^K \lambda_k \boldsymbol{\tau}_k$, where $\lambda_k$ denotes the merging coefficient corresponding to the task vector $\boldsymbol{\tau}_k$. However, existing methods primarily focus on improving the accuracy of ID data,  neglecting the performance on out-of-distribution OOD data.





\begin{figure}[t]
  
  \centering
  
   
   \includegraphics[width=\textwidth]{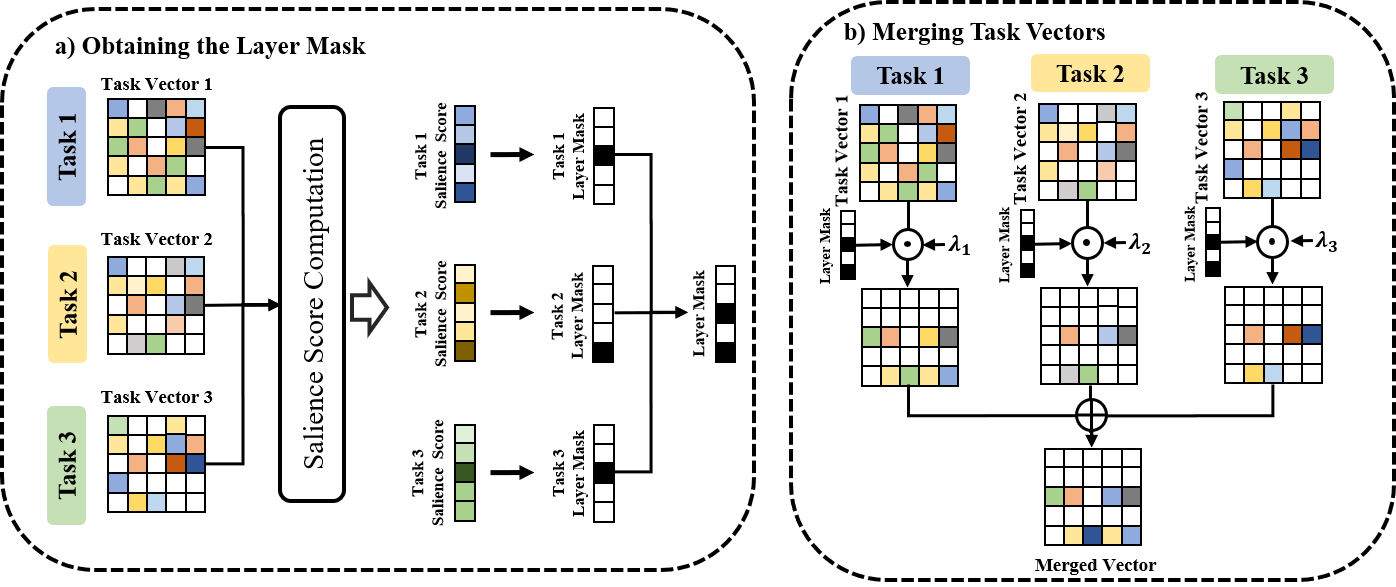}

\caption{\small{{Illustration of our LwPTV  framework: (a) Obtaining the Layer Mask; (b) Merging Task Vectors.}  Each row of the task vector block represents the layer-wise parameter.  A white color denotes a value of 0, whereas a black color signifies a value of 1. The layer-wise salience score of each task vector is calculated, and a threshold operation is performed to generate a mask. The task vector is then layer-pruned through the mask. Finally, these pruned task vectors are integrated into the pre-trained model to form a new merged model.}}
   \label{Overview}
\end{figure}

\vspace{-3mm}
\section{Our proposed method}
\label{sec:Methodology}
\vspace{-2mm}

This work introduces a novel plug and play method, denoted as Layer-wise Pruning Task Vector for model merging (LwPTV),  to improve the generalization capabilities of merging models, whose overview is shown in Fig.\ref{Overview}. We first give the motivation and analysis about the task vector in Sec.\ref{sec:observation} and then introduce the proposed method in Sec.\ref{sec:Promethod}. 


\vspace{-2mm}
\subsection{Motivation and analysis about task vector}
\label{sec:observation}







As shown in Fig.\ref{fig:ID_OOD}, the merged model exhibits inferior generalization capabilities on OOD tasks but superior performance on ID tasks when compared with the pre-trained model. According to Eq.~\ref{Eq3-2}, merged model $\boldsymbol{\theta}_m$ is composed of $\boldsymbol{\theta}_{pre}$ and $\{\boldsymbol{\tau}_k\}_{k=1}^{K}$. The pre-trained model mainly contains valuable parameters for handling OOD tasks as indicated by Fig.\ref{fig:ID_OOD}, and $\boldsymbol{\tau}_k$ have the parameters beneficial for the $k$-th ID task.
Motivated by this observation, our idea to improve OOD generalization for merged models by discarding certain redundant parameter layers within the task vectors and replacing them with the corresponding layers from the pretrained model, making the merged models ``closer'' to pretrained models where necessary. By doing this, we face two fundamental questions:
(1) \emph{\textbf{Which parameters from task vectors should be pruned?}} (2) \emph{\textbf{Can the pruning maintain the performance of merged model on ID tasks?}}

Before answering these questions, we provide a theoretical analysis of the task vector inspired by \cite{LZZC25}, a state-of-the-art theoretical work on the generalization of task vectors. \cite{LZZC25} theoretically analyzes one-layer Transformer models  
 based on discriminative patterns that are unique for different tasks. From Corollary 2 and Lemma 1 of \citep{LZZC25}, one can see that,  some neurons of the task vectors for Transformer models can learn the discriminative patterns, while other neurons cannot. Without loss of generality, our theoretical analysis focuses on Transformer models e.g., ViT \cite{dosovitskiy2020image} and considers one layer, where the task vector in (\ref{Eq3-2}) is simplified to $\boldsymbol{\tau}_k=\boldsymbol{\tau}_k^1$. 
Let $g(\boldsymbol{\tau}_k, \mathcal{S})$ denote the weights of the neuron at the index set $\mathcal{S} \subset \mathcal{L}$ in the task vector $\boldsymbol{\tau}_k$, where $\mathcal{L}$ is the set of all neuron indices. Define the diversity (\emph{DV}) of $\boldsymbol{\tau}_k$ at  neurons from $\mathcal{S}$, with respect to a collection $\{\boldsymbol{\tau}_j\}_{j=1}^K$, as
\begin{equation}\label{diversity}
\text{DV}\left(g(\boldsymbol{\tau}_k, \mathcal{S}); \{\boldsymbol{\tau}_j\}_{j=1}^K\right) = \mathbb{E}  
\Big[  \Big\| g(\boldsymbol{\tau}_k, \mathcal{S}) - \frac{1}{K} \sum_{j=1}^K g(\boldsymbol{\tau}_j, \mathcal{S}) \Big\|  \Big].
\end{equation}
Then, we can derive the following proposition, the proof of which can be found in Appendix \ref{sec:Proposition 1}.

\begin{proposition}\label{prop: condition}
With a high probability, there exists a set of neurons with indices in $\mathcal{S}\in\mathcal{L}$, $|\mathcal{S}|\geq 1$, where all neurons in $\mathcal{S}$ for $\boldsymbol{\tau}_{k_1}$ learn discriminative features, while all neurons in $\mathcal{S}$ for $\boldsymbol{\tau}_{k_2}$ fail to learn discriminative features, $k_1\neq k_2\in[K]$. 
Then, we have
\begin{equation}
    \text{DV}(g(\boldsymbol{\tau}_{k_1}, \mathcal{S});\{\boldsymbol{\tau}_j\}_{j=1}^K)> \sqrt{K}\cdot \Omega(\text{DV}(g(\boldsymbol{\tau}_{k_2}, \mathcal{S});\{\boldsymbol{\tau}_j\}_{j=1}^K))
\end{equation}
\end{proposition}

This result formalizes the intuition that discriminative neurons (parameters) of task vectors exhibit significantly higher variability across tasks compared to non-discriminative ones. It thus provides a theoretical justification for using diversity as a criterion for selective pruning. In particular, parameters with low diversity, which do not contribute to task-specific discriminative features, can be considered redundant or non-informative. Consequently, pruning these low-diversity components from $\boldsymbol{\tau}_k$ and replacing them with their counterparts from the pre-trained model $\boldsymbol{\theta}_{pre}$ is unlikely to degrade performance on ID tasks. Instead, it preserves ID performance while recovering the generalization ability of the merged model by reintroducing OOD-relevant information encoded in $\boldsymbol{\theta}_{pre}$.


\vspace{-2mm}
\subsection{Layer-wise pruning of task vectors}
\label{sec:Promethod}


Based on the aforementioned analysis, we consider a straightforward method to enhance the generalization ability of merging models while preserving its ID  performance:  Pruning the parameters in task vectors with low diversity. Recalling that in the deep neural network model, the information learned by each layer is usually different, and the different layers in each task vector usually have different contributions for the merging model as stated by  \cite{yangadamerging}.   Motivated by Eq. \eqref{diversity}, we introduce a layer-wise salience score:

\vspace{-1.5em}
\begin{eqnarray}
\begin{aligned}
& s_k^l=\mathbb{E}\Big[\Big\|\boldsymbol{\tau}^l_k-\frac{1}{K}\sum_{k=1}^K\boldsymbol{\tau}^l_k\Big\|\Big], s_k^l \in \mathbb{R}_{\geq 0},
\label{Eq3-3}
\end{aligned}
\end{eqnarray}
where  $s_k^l$ represents the salience score of the $l$-th layer of the $\boldsymbol{\tau}_k$. In Eq. \eqref{Eq3-3},  we first compute the difference between $k$-th task vector and mean of all task vectors in layer $l$, which is a $d_l$-dimensional vector with $d_l$ denoting the dimension of $\boldsymbol{\tau}^l_k$; then we compute the absolute average of the vector along the dimension $d_l$, resulting in a scalar value. We use $s_k^l$ to estimate whether the layer-wise parameters contain task-specific information, i.e. have  learned the task-relevant discriminative features. The greater the distance between $k$-th task vector and mean task vectors, the larger the 
$s_k^l$, indicating that the 
$l$-th layer of the 
$k$-th fine-tuned model contains more task-specific local information. On the contrary, if the value is smaller, it means that the $k$-th task is close to all tasks in current layer. 
That is to say, the parameters of the current layer are redundant, failing to learn the task-relevant discriminative features, so the parameters in this layer can be pruned.


To solve the first problem discussed above, we denote the mask vector as a $L$-dimensional vector, i.e., $ \boldsymbol{m}_k =\{m_k^l\}_{l=1}^{L} \in \mathbb{R}^L $ for the $k$-th task and we design  it according to saliency scores:
\begin{eqnarray}
m_k^l=\left\{ \begin{aligned}
    &1, \:\:\: if \:\:\:s_k^l > \text{sorted}(\mathbf{s}_k)[\lfloor L \cdot \eta \rfloor], \\
    &0, \:\:\: \text{otherwise},
\end{aligned}
\right.
\label{Eq3-4}
\end{eqnarray}
where $\eta$ is the hyperparameter controlling the pruning ratio, and $\text{sorted}(\mathbf{s}_k)[\lfloor L \cdot \eta \rfloor]$ represents 
the $\eta L$-th smallest salience score, ensuring that only the top $1-\eta$ fraction of layers with the highest salience scores are retained.
In other words, for each task vector, layers whose salience scores significantly larger are retained, while layers with lower salience scores—indicating redundant parameters—are replaced with the corresponding pre-trained parameters. This adaptive pruning mechanism ensures that the merged model maintains task-specific adaptations while leveraging the generalization capabilities of the pre-trained model, thereby enhancing OOD robustness.
Considering the mask defined independently for each task vector, there remains a risk that some task-specific mask vectors may remove parameter layers that are crucial for their respective tasks. To solve the second question mentioned earlier, i.e, maintaining the ability of the merged model on ID tasks as much as possible,  we introduce a shared mask vector that consolidates information across all mask vectors:
\begin{eqnarray}
   \hat{\mathbf{m}}= \mathbf{m}_1 \lor \mathbf{m}_2\lor \dots \lor \mathbf{m}_K,
\label{Eq3-6}
\end{eqnarray}
where $\hat{\mathbf{m}}_k \in \mathbb{R}^L$ and $\lor$ is the OR  operation. That is to say, a layer $l$ in the merged model will be pruned only if all task-specific mask vectors agree to remove it, i.e., $\hat{m}_k^l=0$ if and only if $m_{1:K}^l=0$. This ensures that if at least one task requires a particular layer to be retained, we remain it intact in the merged model, thereby preserving critical task-specific information.

Our proposed masking strategy is flexible so that it can be integrated with existing task vector-based merging methods. Specifically, it can be applied to the Task Arithmetic, where the merging model $\hat{\boldsymbol{\theta}} $ is formulated as follows:
\begin{equation} \hat{\boldsymbol{\theta}}_m=\boldsymbol{\theta}_{pre}+\hat{\mathbf{m}} \odot \sum_{k=1}^K \lambda_k \boldsymbol{\tau}_k.
\label{Eq3-5}
\end{equation}

 In which, the $\{\lambda_k\}_{k=1}^K$ values are identical.
 Besides, this mask vector can also be applied to other methods. For example, for AdaMerging series methods, which are designed to optimize coefficient $\lambda_k$ or $\lambda_k^l$ with entropy minimization, we can optimize $\lambda_k$ or $\lambda_k^l$ based on \eqref{Eq3-5}. Considering the learnable $\lambda_k^l$ will be  amplified when we introducing the mask, we can further use   $\hat{\lambda}^l_k=\eta \cdot \lambda^l_k$ to scale the coefficient after optimizing the \eqref{Eq3-5}. 
 A detailed analysis is in the \ref{sub:analysis_eta} of Appendix.
 We defer the workflow when combining ours with existing methods to Algorithm \ref{sub:algorithmic2} of Appendix.

\begin{rmk}

By combining the definition of saliency score in (\ref{Eq3-3}), Proposition \ref{prop: condition} indicates that parameters that learn task-specific discriminative patterns result in a large saliency score because these parameters are diverse with a large variance. Then, that layer in the merged model will be maintained because it preserves task-specific information. In contrast, parameters that cannot learn task-specific discriminative patterns are close to each other, leading to a small saliency score. Then, that layer in the merged model will be pruned.  This theoretically explains the success of the proposed layer-wise pruning algorithm, which only maintains layers that learn discriminative patterns.
\end{rmk}

\begin{table}[t]
  \caption{\small{Summarization of model merging methods, where we consider following categories: ``No Training/ Test'' means no need Training/Test set in  designing merging coefficient or merging models, ``Pruning'' indicates discarding parameters in task vectors, ``orthogonal'' in merging coefficients indicates that method is orthogonal to  merging coefficient designed by others, ``Designed for OOD'' indicates that the method is specifically designed for merging model on OOD, ``Available on OOD'' indicates that the merging model is available to OOD data.}}
 \scriptsize
  \label{OOD}
  \centering
  \scriptsize
  \setlength{\tabcolsep}{3pt}
  \resizebox{\textwidth}{!}{
  \begin{tabular}{l|cccccccc}

   \toprule \rowcolor{gray!30}  Method  &No Training set&No Test&    Pruning    & Merging coefficients & Available on OOD &Designed for OOD &Storage cost \\    
    \midrule
    Weight Averaging \cite{wortsman2022model}  &\cmark  & \cmark &  \xmark    &    Uniformed   &\cmark &\xmark & high \\
    Fisher Merging \cite{matena2022merging} &\xmark   & \cmark &  \xmark    &    Fisher Matric  &\cmark &\xmark & high\\
     RegMean \cite{jindataless}  &\xmark  &\cmark &  \xmark    &    Inner Product Matric  & \cmark &\xmark & high\\
     Task Arithmetic \cite{ilharco2022editing} & \cmark & \cmark  &  \xmark    &    Uniformed    & \cmark &\xmark & high \\
      Ties-Merging \cite{yadav2023resolving}  & \cmark & \cmark  &  Magnitude  Metric   &    Uniformed  & \cmark&\xmark & low \\
       AdaMerging \cite{yangadamerging} &  \cmark &  \xmark &  \xmark    &    Optimized & \cmark &\xmark & high\\

     Surgery \cite{yangrepresentation} &  \cmark &  \xmark &  \xmark    &  orthogonal &   \xmark &\xmark & high \\
       




      PCB-MERGING \cite{du2024parameter} &  \cmark & \cmark  & PCB matrix     &   $ (\hat{\boldsymbol{\beta}}_k \odot\lambda_k) /\sum_{k=1}^K\hat{\boldsymbol{\beta}}_i$&   \cmark  &\xmark& low \\  
      \method (Ours)  &  \cmark & \cmark  & Salience Score    &   orthogonal &   \cmark &\cmark & low \\  
    \bottomrule
  \end{tabular}
  }
  \vspace{-15pt}
\end{table}

\vspace{-2mm}
\section{Experiments}
\label{sec:Experiments}
\vspace{-2mm}


\textbf{Datasets.} Following
\cite{ilharco2022editing}, \cite{yadav2023resolving} and \cite{yangadamerging}, we perform model merging on eight image classification datasets, including  Cars \cite{krause20133d}, DTD \cite{cimpoi2014describing}, EuroSAT \cite{helber2019eurosat}, GTSRB \cite{stallkamp2011german}, MNIST \cite{deng2012mnist}, RESISC45 \cite{cheng2017remote}, SUN397 \cite{xiao2016sun}, and SVHN \cite{netzer2011reading}. To evaluate the generalization capability of the merged model on OOD datasets, we have selected a collection of 13 image classification datasets to serve as OOD benchmarks for our experiments \cite{wanglocalizing, zhou2022learning}. 
The OOD benchmarks comprises the following datasets: CIFAR10 (C10) \cite{krizhevsky2009learning}, CIFAR100 (C100) \cite{krizhevsky2009learning}, EMNIST (EMN) \cite{cohen2017emnist}, FashionMNIST (FMN)\cite{xiao2017fashion}, FER2013 (FER) \cite{goodfellow2013challenges}, Flowers102 (F102) \cite{nilsback2008automated}, Food101 (F101) \cite{bossard2014food},  KMNIST \cite{clanuwat2018deep} (KMN), OxfordIIITPet (OxP) \cite{parkhi2012cats}, PCAM (PCA) \cite{veeling2018rotation},     RenderedSST2 (RSS) \cite{socher2013recursive, radford2019language}, STL10 (STL) \cite{coates2011analysis}  and ImageNet-1k (Ima) \cite{deng2009imagenet}. More details about  datasets are available in Appendix \ref{sub:datasets}.

\textbf{Baselines \& details.} We consider  following baselines, including Weight Averaging \cite{wortsman2022model}, Fisher Merging \cite{matena2022merging}, RegMean \cite{jindataless}, Task Arithmetic \cite{ilharco2022editing}, Ties-Merging \cite{yadav2023resolving}, AdaMerging \cite{yangadamerging}, Surgery \cite{yangrepresentation} and PCB-MERGING \cite{du2024parameter}. We employ ViT-B/32, ViT-L/14 and ViT-H/14 from CLIP \cite{DBLP:conf/icml/RadfordKHRGASAM21} as our pre-trained models, repsectively. Motivated by \cite{zhumodel}, we employ the H-score as  evaluation metric, defined as the harmonic mean of the performance on ID and OOD datasets and stated in Appendix \ref{sub:Evaluation_Metrics}. Consistent with previous studies \cite{yadav2023resolving,yangadamerging, yangrepresentation, du2024parameter}, we use the following parameters: For Task Arithmetic, we set $\lambda=0.3$; for Ties, we set $\lambda=0.3$ for ViT-B/32, $\lambda=0.4$ for ViT-L/14, and $\lambda=0.4$ for ViT-H/14, respectively, and retain the top-$20\%$ parameters for all architectures; for PCB, we set $\lambda=1.2$ and mask ratio $r=0.05$. For our LwPTV, we set pruning ratio $\eta=0.7$ for all experiments.




\vspace{-2mm}
\subsection{Main results}
\textbf{Comparison with other methods.} We summarize the characteristics of baselines and ours in Tab.~\ref{OOD} for a clear comparison. 
As listed in Tab.\ref{ViT-B-32}, we evaluate the performance of merging model ViT-B/32 and ViT-L/14 on  ID  and  OOD datasets respectively. Individual  indicates $K$ task-specific fine-tuned model parameters $ \{\boldsymbol{\theta}_{k}\}^K_{k=1}$ without model merging. For ID tasks, each task is only tested by the corresponding individual model; for OOD tasks, each OOD task can be evaluated by all individual models, respectively, followed by the average operation. 
Due to the unavailability of Fisher Merging and RegMean, and their suboptimal performance on ID datasets, we did not test their performance on OOD datasets. Besides, Surgey is not available in OOD tasks since it needs to learn a representation surgey module on ID task. The detailed performance of various methods including Fisher Merging, RegMean and Surgey on the ID dataset are deferred to Appendix \ref{sec:ID_Per} due to limited space. 

From Tab.\ref{ViT-B-32}, we have the following observations: (1) For the non-model merging baseline, the Pretrained model exhibits robust generalization capabilities on OOD datasets, whereas the Individual fine-tuned models demonstrates  inferior OOD performance and superior ID performance.  The underlying cause  is that the Pretrained model acquires generalized features from the large-scale pre-training datasets and the Individual Fine-tuned model undergoes feature distortion specific to the current task. (2) For the model merging baselines, Weight Averaging is equivalent to computing the mean of various task vectors and incorporating these into the pretrained model. This method minimizes alterations to the parameters of the pre-trained model, thereby yielding superior OOD generalization performance. The Ties method prunes task vectors. Consequently, its OOD generalization performance is also commendable. However, their performance on ID datasets is relatively poor.
TW AdaMerging, LW AdaMerging, and PCB respectively allocate task-level, layer-level, and parameter-level coefficients to the task vector. This modification drastically alters the parameters of the pre-trained model, which thus enhances ID datasets but compromises the model’s generalization capacity on OOD datasets. (3) \method  
 is a plug-and-play technique that enhances the generalization of existing model merging schemes on OOD datasets while minimizing the degradation on (even improving) the ID performance. For instance, on ViT-B/32, ours yields additional improvements of $7.4\%$, $1.9\%$, $1.5\%$, and $4.9\%$ in H-score values compared to the baseline results of Task Arithmetic, Ties, LW AdaMerging, and PCB, respectively. 
 (4) 
 To investigate whether ours is effective for more larger pre-trained model, we conducted  experiments on ViT-H/14 in Fig.\ref{fig:ViT-H-14-per}, proving its effectiveness and robustness on pre-trained model; see more details in Appendix \ref{sub:ViT-H-14}. 
\vspace{-10pt}

\begin{table}[t]
\centering
\caption{Performance comparison for ID and OOD datasets on ViT-B/32 and ViT-L/14.}
\label{ViT-B-32}
\small
\setlength{\tabcolsep}{0.5mm}
\resizebox{\textwidth}{!}{%
\begin{tabular}{@{}cccccccccccccccccc@{}}
\toprule
\multicolumn{2}{c}{\textbf{Dataset (→)}}  &  \multicolumn{1}{c}{\textbf{In Domain}} & \multicolumn{14}{c}{\textbf{Out of Domain}} &\multirow{2}{*}{ \textbf{H-score}} \\ 
  \cmidrule(lr){3-3} \cmidrule(lr){4-17}  
 \multicolumn{2}{c}{\textbf{Method (↓)}} & \textbf{Avg.} & \textbf{C10} & \textbf{C100} & \textbf{EMN} & \textbf{FMN} & \textbf{FER}& \textbf{F102}& \textbf{F101}& \textbf{KMN}& \textbf{OxP} & \textbf{PCA} & \textbf{RSS} & \textbf{STL}& \textbf{Ima} & \textbf{Avg.}  \\ 
\midrule
\multirow{17}{*}{\raisebox{-15mm}{\rotatebox{90}{\textbf{ViT-B/32}}}}
&Pretrained &48.0 & 89.8 & 64.2 & 17.2 &63.0 &39.0& 66.3 &82.7 &9.8 & 87.4 & 60.6 &58.6 & 97.1 & 63.3 & 61.5& 53.9  \\ 

&Individual & 90.5 & 65.5 & 37.3 & 18.2 & 54.8 & 32.2& 47.8& 58.3 &8.5 & 76.9 & 54.1 & 55.9 & 88.4 &48.5 &49.7& 64.2   \\ 


\cmidrule(lr){2-18} 
&Weight Averaging \cite{wortsman2022model}
& 65.8 & 89.5 &64.2 & 28.8 & 65.1& 36.4&64.5 &79.7& 7.4&86.7 &59.4 & 57.7 & 96.4 &61.6 &61.3& 63.5  \\ 



\cmidrule(lr){2-18} 
&Task Arithmetic \cite{ilharco2022editing} 
& 69.1 & 76.3 & 41.9 & 28.5 & 63.9 & 26.3 & 49.4 & 55.8 & \textbf{9.0} & 75.4 & 54.0 & 53.4 & 87.8 & 45.0 
& 51.3 & 58.9 \\
&\cellcolor{gray!10}\textbf{w/ \method} (Ours)&\cellcolor{gray!10}\textbf{72.8} & \cellcolor{gray!10}\textbf{88.3} & \cellcolor{gray!10}\textbf{60.1} & \cellcolor{gray!10}\textbf{30.4} & \cellcolor{gray!10}\textbf{65.9} & \cellcolor{gray!10}\textbf{36.5} & \cellcolor{gray!10}\textbf{62.9} & \cellcolor{gray!10}\textbf{79.9} & \cellcolor{gray!10}7.4 & \cellcolor{gray!10}\textbf{85.4} & \cellcolor{gray!10}\textbf{59.5} & \cellcolor{gray!10}\textbf{59.0} & \cellcolor{gray!10}\textbf{96.4} & \cellcolor{gray!10}\textbf{60.6} & \cellcolor{gray!10}\textbf{60.9}\textsuperscript{\best{+9.6}} & \cellcolor{gray!10}\textbf{66.3}\textsuperscript{\best{+7.4}} \\
\cmidrule(lr){2-18}

&Ties \cite{yadav2023resolving}&\textbf{72.9}&85.3&55.2&26.5&64.1&33.6&59.0&74.0&\textbf{8.5}&83.0&58.4&55.6&94.4&56.3&58.0&64.6\\ 
&\cellcolor{gray!10}\textbf{w/ \method} (Ours)&\cellcolor{gray!10}72.7&\cellcolor{gray!10}\textbf{89.1}&\cellcolor{gray!10}\textbf{61.5}&\cellcolor{gray!10}\textbf{29.3}&\cellcolor{gray!10}\textbf{65.7}&\cellcolor{gray!10}\textbf{37.0}&\cellcolor{gray!10}\textbf{63.5}&\cellcolor{gray!10}\textbf{80.4}&\cellcolor{gray!10}7.4&\cellcolor{gray!10}\textbf{85.5}&\cellcolor{gray!10}\textbf{59.8}&\cellcolor{gray!10}\textbf{59.4}&\cellcolor{gray!10}\textbf{96.6}&\cellcolor{gray!10}\textbf{61.1}&\cellcolor{gray!10}\textbf{61.3}\textsuperscript{\best{+3.3}}&\cellcolor{gray!10}\textbf{66.5}\textsuperscript{\best{+1.9}}\\
\cmidrule(lr){2-18}

&TW AdaMerging \cite{yangadamerging} & 71.1 &78.6 &45.9 &30.6&\textbf{63.5} &28.7& 51.1& 59.9&\textbf{8.2}&79.0 & 53.4 & 54.0 & 89.1 & 49.5 &53.2&60.9  \\ 
 & \cellcolor{gray!10}\textbf{w/ \method} (Ours)&\cellcolor{gray!10}\textbf{75.4} &\cellcolor{gray!10}\textbf{81.2}&\cellcolor{gray!10}\textbf{47.0}&\cellcolor{gray!10}\textbf{31.3}&\cellcolor{gray!10}62.6&\cellcolor{gray!10}\textbf{32.3}&\cellcolor{gray!10}\textbf{56.3}&\cellcolor{gray!10}\textbf{73.1}&\cellcolor{gray!10}7.7 &\cellcolor{gray!10}\textbf{81.2} &\cellcolor{gray!10}\textbf{55.7} &\cellcolor{gray!10}\textbf{57.8} &\cellcolor{gray!10}\textbf{93.7} &\cellcolor{gray!10}\textbf{54.1}  &\cellcolor{gray!10}\textbf{56.5}\textsuperscript{\best{+3.3}} &\cellcolor{gray!10}\textbf{64.6}\textsuperscript{\best{+3.7}}  \\
\cmidrule(lr){2-18}

&TW AdaMerging++ \cite{yangadamerging} &73.7 &79.3 &45.5 &28.1&\textbf{63.2} &30.4& 51.7& 64.3&\textbf{8.3}&78.1 & 54.6 & 54.1 & 90.6 & 50.6 &53.7&62.1  \\ 
 &\cellcolor{gray!10} \textbf{w/ \method} (Ours)&\cellcolor{gray!10}\textbf{76.1} &\cellcolor{gray!10}\textbf{83.2}&\cellcolor{gray!10}\textbf{49.1}&\cellcolor{gray!10}\textbf{30.7}&\cellcolor{gray!10}62.8 &\cellcolor{gray!10}\textbf{32.5}&\cellcolor{gray!10}\textbf{55.4}&\cellcolor{gray!10}\textbf{73.8}&\cellcolor{gray!10}8.0 &\cellcolor{gray!10}\textbf{79.9} &\cellcolor{gray!10}\textbf{56.3} &\cellcolor{gray!10}\textbf{56.7} &\cellcolor{gray!10}\textbf{94.2}&\cellcolor{gray!10}\textbf{53.5}   &\cellcolor{gray!10}\textbf{56.6}\textsuperscript{\best{+2.9}}&\cellcolor{gray!10}\textbf{64.9}\textsuperscript{\best{+2.8}} \\
\cmidrule(lr){2-18}

&LW AdaMerging \cite{yangadamerging} &\textbf{80.1} &82.1 & 50.7 & 28.7 &63.2& 37.8 & 58.3 &73.8 &\textbf{8.8}& 82.4 &  57.3 & 58.0 & 94.3& 54.6&57.7&67.1  \\ 
 &\cellcolor{gray!10}\textbf{w/ \method} (Ours)&\cellcolor{gray!10}77.3 &\cellcolor{gray!10}\textbf{88.8}&\cellcolor{gray!10}\textbf{61.3}&\cellcolor{gray!10}\textbf{30.6}&\cellcolor{gray!10}\textbf{66.1}&\cellcolor{gray!10}\textbf{38.4}&\cellcolor{gray!10}\textbf{64.3}&\cellcolor{gray!10}\textbf{80.6}& \cellcolor{gray!10}7.6 &\cellcolor{gray!10}\textbf{86.3} &\cellcolor{gray!10}\textbf{60.1} &\cellcolor{gray!10}\textbf{59.4} &\cellcolor{gray!10}\textbf{96.6} &\cellcolor{gray!10}\textbf{61.2}  &\cellcolor{gray!10}\textbf{61.7}\textsuperscript{\best{+4.0}} &\cellcolor{gray!10}\textbf{68.6}\textsuperscript{\best{+1.5}}\\
\cmidrule(lr){2-18}

&LW AdaMerging++ \cite{yangadamerging} &\textbf{81.1} &83.9 & 51.7  &27.2  &62.7 & 36.1 & 57.4 &74.2 &\textbf{9.1}& 82.4 &  56.4 & 57.8 & 94.3& 54.7&57.6&67.4  \\ 
 & \cellcolor{gray!10}\textbf{w/ \method} (Ours)& \cellcolor{gray!10}77.1 &\cellcolor{gray!10}\textbf{89.0}&\cellcolor{gray!10}\textbf{61.8}&\cellcolor{gray!10}\textbf{29.9}&\cellcolor{gray!10}\textbf{65.8}&\cellcolor{gray!10}\textbf{37.5}&\cellcolor{gray!10}\textbf{63.9}&\cellcolor{gray!10}\textbf{80.4}&\cellcolor{gray!10}7.8 &\cellcolor{gray!10}\textbf{86.0} &\cellcolor{gray!10}\textbf{59.6} &\cellcolor{gray!10}\textbf{58.9} &\cellcolor{gray!10}\textbf{96.5} &\cellcolor{gray!10}\textbf{61.1}  &\cellcolor{gray!10}\textbf{61.4}\textsuperscript{\best{+3.8}}&\cellcolor{gray!10}\textbf{68.4}\textsuperscript{\best{+1.0}} \\
\cmidrule(lr){2-18}

&PCB-MERGING \cite{du2024parameter} & 75.8  &77.7& 43.6&  26.0&61.2 &29.4& 52.9& 65.4& \textbf{8.8} &79.3 & 53.6 &52.2  &90.5  & 50.8&53.2 &62.5   \\
 & \cellcolor{gray!10}\textbf{w/ \method} (Ours)&\cellcolor{gray!10}\textbf{77.3} & \cellcolor{gray!10}\textbf{86.5}&\cellcolor{gray!10}\textbf{57.1} &\cellcolor{gray!10}\textbf{29.8} &\cellcolor{gray!10}\textbf{64.3}  &\cellcolor{gray!10}\textbf{35.9}  &\cellcolor{gray!10}\textbf{60.7}&\cellcolor{gray!10}\textbf{78.2}  &\cellcolor{gray!10}7.6 &\cellcolor{gray!10}\textbf{84.6} &\cellcolor{gray!10}\textbf{58.4} &\cellcolor{gray!10}\textbf{57.7} &\cellcolor{gray!10}\textbf{95.6} &\cellcolor{gray!10}\textbf{59.5}  &\cellcolor{gray!10}\textbf{59.7}\textsuperscript{\best{+6.5}} &\cellcolor{gray!10}\textbf{67.4}\textsuperscript{ \best{+4.9}} \\


\bottomrule
\\[-1.5ex]
\multirow{16}{*}{\raisebox{-15mm}{\rotatebox{90}{\textbf{ViT-L/14}}}}
&Pretrained &64.5 & 95.6 &75.8 & 15.6 &66.9 &38.2& 79.2 &93.1 &10.4 & 93.4 & 51.2 &68.9 & 99.4 & 75.5& 66.4& 65.4 \\ 

&Individual& 94.2 & 88.6 & 65.2 & 19.3 & 66.6 & 38.5 & 73.7 & 89.4 &10.2 & 92.4 & 59.4 & 62.4 & 97.8 & 72.5&64.3& 76.4  \\ 


\cmidrule(lr){2-18}
&Weight Averaging \cite{wortsman2022model} & 79.6 & 95.9 & 79.6 & 22.2 & 71.5& 38.7& 79.0 &92.4& 8.7& 93.6 & 59.0 & 64.1 & 99.2 & 75.8 & 67.7& 73.2 \\ 


\cmidrule(lr){2-18}

&Task Arithmetic \cite{ilharco2022editing} & 84.5&  93.4 & 73.0 &  \textbf{28.7} &  73.2 & 38.2 & 74.2 &  88.6 & \textbf{9.3} & 92.7 &\textbf{61.4} & 58.6 & 98.2& 72.9 &66.3& 74.3 \\ 
 & \cellcolor{gray!10}\textbf{w/ \method} (Ours) &\cellcolor{gray!10}\textbf{85.6}&\cellcolor{gray!10}\textbf{95.6}&\cellcolor{gray!10}\textbf{78.8}&\cellcolor{gray!10}24.4&\cellcolor{gray!10}\textbf{73.4} &\cellcolor{gray!10}\textbf{40.1}&\cellcolor{gray!10}\textbf{77.6}&\cellcolor{gray!10}\textbf{92.5}&\cellcolor{gray!10}8.4
&\cellcolor{gray!10}\textbf{93.8 }&\cellcolor{gray!10}60.2 &\cellcolor{gray!10}\textbf{63.4} &\cellcolor{gray!10}\textbf{99.1}&\cellcolor{gray!10}\textbf{76.0}  &\cellcolor{gray!10}\textbf{67.9}\textsuperscript{\best{+1.6}}&\cellcolor{gray!10}\textbf{75.7}\textsuperscript{\best{+1.4}}  \\ 
\cmidrule(lr){2-18}

&Ties \cite{yadav2023resolving} & 86.0 & 94.5  &74.6  &\textbf{27.2}& 72.5& 38.8& 75.1& 90.3 &8.5& 92.9&\textbf{60.5} & 59.7 &98.5&  74.2  &66.7& 75.1  \\ 
 &\cellcolor{gray!10} \textbf{w/ \method} (Ours) &\cellcolor{gray!10}\textbf{86.9 }&\cellcolor{gray!10}\textbf{95.1}&\cellcolor{gray!10}\textbf{76.9}&\cellcolor{gray!10}26.2&\cellcolor{gray!10}\textbf{73.2}&\cellcolor{gray!10}\textbf{39.6}&\cellcolor{gray!10}\textbf{76.5}  &\cellcolor{gray!10}\textbf{92.3}&\cellcolor{gray!10}\textbf{8.7}
&\cellcolor{gray!10}\textbf{93.7} &\cellcolor{gray!10}59.5 &\cellcolor{gray!10}\textbf{63.7} &\cellcolor{gray!10}\textbf{99.0}&\cellcolor{gray!10}\textbf{75.7}  &\cellcolor{gray!10}\textbf{67.7}\textsuperscript{\best{+1.0}}&\cellcolor{gray!10}\textbf{76.1}\textsuperscript{ \best{+1.0}}  \\ 
\cmidrule(lr){2-18}

&TW AdaMerging \cite{yangadamerging} &84.3 &\textbf{90.9}&\textbf{67.1}&25.2& \textbf{71.9} &36.6& \textbf{69.3}& 85.2& 9.2& 91.8 & \textbf{59.4} &57.2 &97.2&  69.6 &63.9& 72.7  \\ 
 &\cellcolor{gray!10} \textbf{w/ \method} (Ours) &\cellcolor{gray!10}\textbf{89.0}&\cellcolor{gray!10}90.4&\cellcolor{gray!10}65.4&\cellcolor{gray!10}\textbf{25.4}&\cellcolor{gray!10}70.7 &\cellcolor{gray!10}\textbf{39.7}&\cellcolor{gray!10}68.5&\cellcolor{gray!10}\textbf{87.5}&\cellcolor{gray!10}\textbf{9.2}&\cellcolor{gray!10}\textbf{92.7} &\cellcolor{gray!10}58.6 &\cellcolor{gray!10}\textbf{63.2} & \cellcolor{gray!10}\textbf{97.3}&\cellcolor{gray!10}\textbf{70.6} &\cellcolor{gray!10}\textbf{64.5}\textsuperscript{\best{+0.6}}&\cellcolor{gray!10}\textbf{74.8}\textsuperscript{ \best{+2.1}}  \\ 
\cmidrule(lr){2-18}
&TW AdaMerging++ \cite{yangadamerging} &87.5 &91.3 &66.7 &24.5& \textbf{71.0} &36.9& \textbf{69.1}& 86.7& 8.2& 91.8 & 57.5 &57.6 &97.5&  70.2  & 63.7& 73.7  \\ 
 &\cellcolor{gray!10} \textbf{w/ \method} (Ours) &\cellcolor{gray!10}\textbf{89.7}&\cellcolor{gray!10}\textbf{91.3}&\cellcolor{gray!10}\textbf{67.0}&\cellcolor{gray!10}\textbf{25.8}&\cellcolor{gray!10}70.6 &\cellcolor{gray!10}\textbf{39.2}& \cellcolor{gray!10}68.6&\cellcolor{gray!10}\textbf{88.4}& \cellcolor{gray!10}\textbf{8.4}&\cellcolor{gray!10}\textbf{92.6} &\cellcolor{gray!10}\textbf{57.8} &\cellcolor{gray!10}\textbf{65.3} &\cellcolor{gray!10}\textbf{97.6}&\cellcolor{gray!10}\textbf{71.2}&\cellcolor{gray!10}\textbf{64.9}\textsuperscript{\best{+1.2}}&\cellcolor{gray!10}\textbf{75.3}\textsuperscript{\best{+1.6}}  \\ 
\cmidrule(lr){2-18}
&LW AdaMerging \cite{yangadamerging} &\textbf{90.8}  &92.2 & 69.5 &\textbf{29.9}  &72.1 & 38.5 & 70.8 &90.1 &8.6& 92.6 & \textbf{60.1}&63.3 &98.0&  72.5 &66.0& 76.4\\ 
 & \cellcolor{gray!10}\textbf{w/ \method} (Ours)  &\cellcolor{gray!10}89.8 &\cellcolor{gray!10}\textbf{95.1}&\cellcolor{gray!10}\textbf{77.0}&\cellcolor{gray!10}26.2&\cellcolor{gray!10}\textbf{73.3}&\cellcolor{gray!10}\textbf{40.4}&\cellcolor{gray!10}\textbf{75.2}&\cellcolor{gray!10}\textbf{92.1}&\cellcolor{gray!10}\textbf{8.7}&\cellcolor{gray!10}\textbf{93.6}&\cellcolor{gray!10}59.9&\cellcolor{gray!10}\textbf{65.5} &\cellcolor{gray!10}\textbf{98.9}&\cellcolor{gray!10}\textbf{75.4}   &\cellcolor{gray!10}\textbf{68.4}\textsuperscript{\best{+2.4}}&\cellcolor{gray!10}\textbf{77.7}\textsuperscript{\best{+1.3}}   \\ 
\cmidrule(lr){2-18}

&LW AdaMerging++ \cite{yangadamerging} &\textbf{91.0}  &93.5 & 71.8 &\textbf{29.9}  &72.3&38.4 & 71.2 &90.3 &8.7&92.8 & \textbf{60.6} & 60.4 & 98.2&  72.7  &66.2& 76.6\\ 
 &\cellcolor{gray!10} \textbf{ w/ \method} (Ours) &\cellcolor{gray!10}89.7 &\cellcolor{gray!10}\textbf{95.1}&\cellcolor{gray!10}\textbf{77.3}&\cellcolor{gray!10}28.0&\cellcolor{gray!10} \textbf{73.1}&\cellcolor{gray!10}\textbf{40.1}&\cellcolor{gray!10}\textbf{75.0}&\cellcolor{gray!10}\textbf{92.1}&\cellcolor{gray!10}\textbf{8.7}&\cellcolor{gray!10}\textbf{93.6}&\cellcolor{gray!10}59.3 &\cellcolor{gray!10}\textbf{65.7} &\cellcolor{gray!10}\textbf{99.0} & \cellcolor{gray!10}\textbf{75.3}   &\cellcolor{gray!10}\textbf{67.9}\textsuperscript{\best{+1.7}}&\cellcolor{gray!10}\textbf{77.3}\textsuperscript{\best{+0.7}}  \\ 
\cmidrule(lr){2-18}

&PCB-MERGING \cite{du2024parameter} & 87.6&93.0  &70.9&\textbf{27.6}& 72.3&38.8 &74.0& 89.6& \textbf{9.1}& 92.7 & \textbf{61.1}& 57.8 &  98.3& 73.5   &66.0 &75.3    \\ 
 & \cellcolor{gray!10}\textbf{w/ \method} (Ours) &\cellcolor{gray!10}\textbf{88.1} &\cellcolor{gray!10}\textbf{94.6}&\cellcolor{gray!10}\textbf{75.3} &\cellcolor{gray!10}27.5 &\cellcolor{gray!10}\textbf{73.0}  &\cellcolor{gray!10}\textbf{39.8}  &\cellcolor{gray!10}\textbf{76.2}& \cellcolor{gray!10}\textbf{91.9}  &\cellcolor{gray!10}8.8&\cellcolor{gray!10}\textbf{93.5}&\cellcolor{gray!10}60.1&\cellcolor{gray!10}\textbf{62.6}&\cellcolor{gray!10}\textbf{98.9}&\cellcolor{gray!10}\textbf{75.5}   &\cellcolor{gray!10}\textbf{67.5}\textsuperscript{\best{+1.5}} &\cellcolor{gray!10}\textbf{76.4}\textsuperscript{ \best{+1.1}}\\

\bottomrule
\end{tabular}
}
\end{table}

\begin{figure}[htbp!]
  \centering
  \begin{minipage}[t]{0.32\textwidth}
    \centering
    \includegraphics[width=\linewidth]{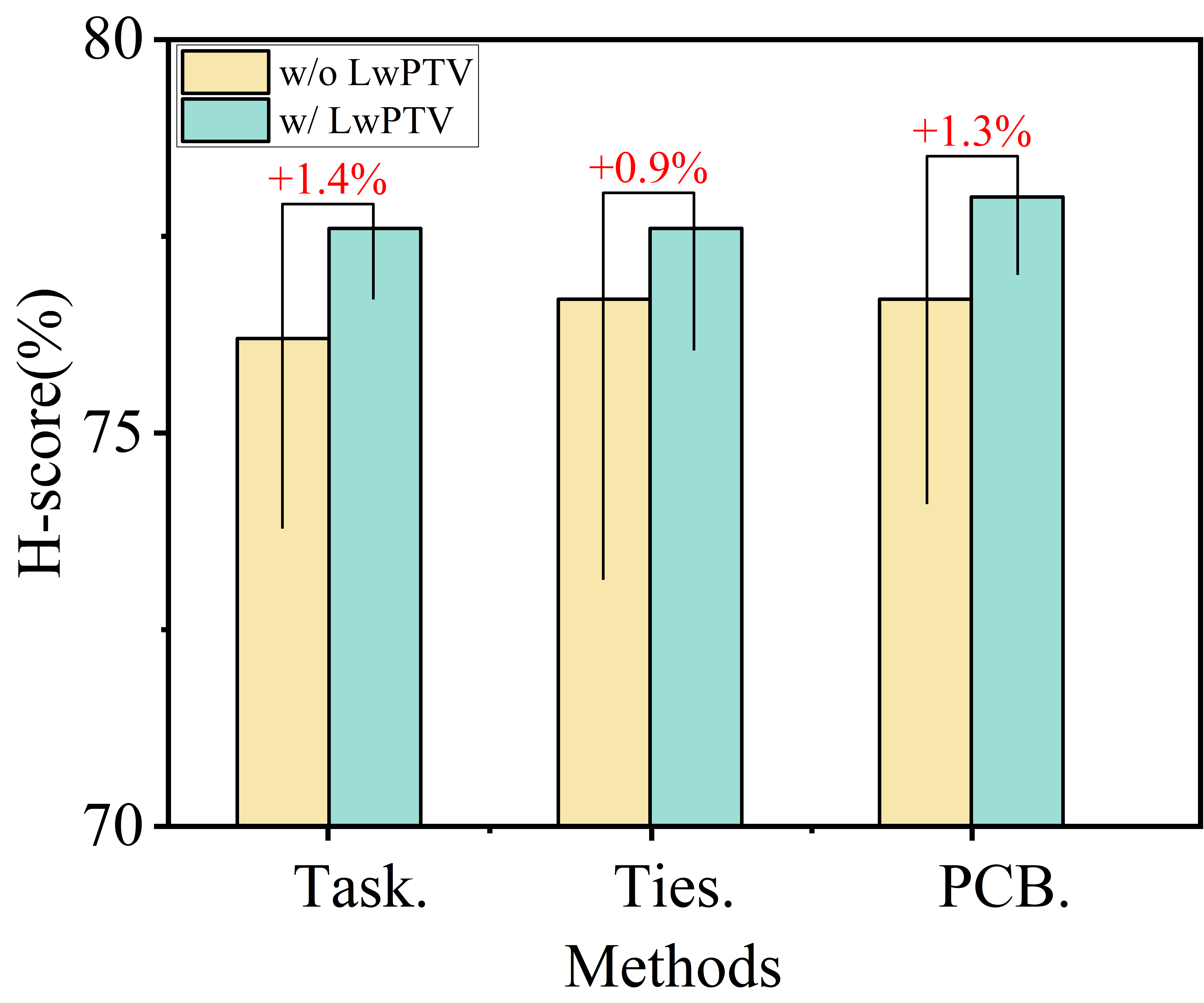}
     \vspace{-1.5em}
    \caption{\small{ H-score of existing model merging methods with and without ours on ViT-H/14.}}
        \vspace{-20pt} 

    \label{fig:ViT-H-14-per}
  \end{minipage}
   \hfill  
  \begin{minipage}[t]{0.32\textwidth}
    \centering
    \includegraphics[width=\linewidth]{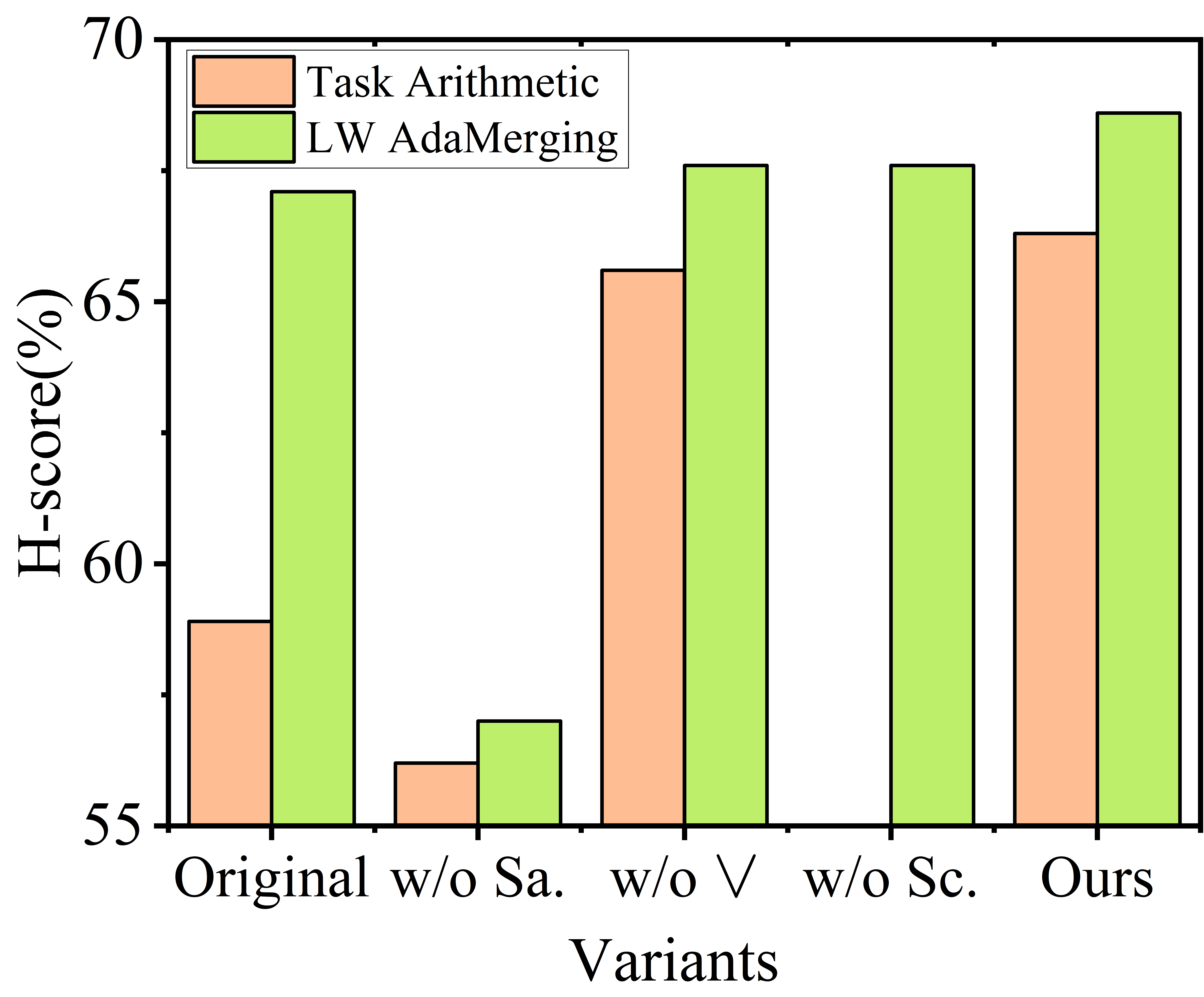}
     \vspace{-1.5em}
    \caption{\small{Ablation study  on ViT-B/32, where Sa, $\lor$ and Sc denotes Salience score, OR operation and scaling, respectively.}}
\label{fig:ablation_study}
  \end{minipage}
  \hfill
  \begin{minipage}[t]{0.32\textwidth}
    \centering
   \includegraphics[width=\linewidth]{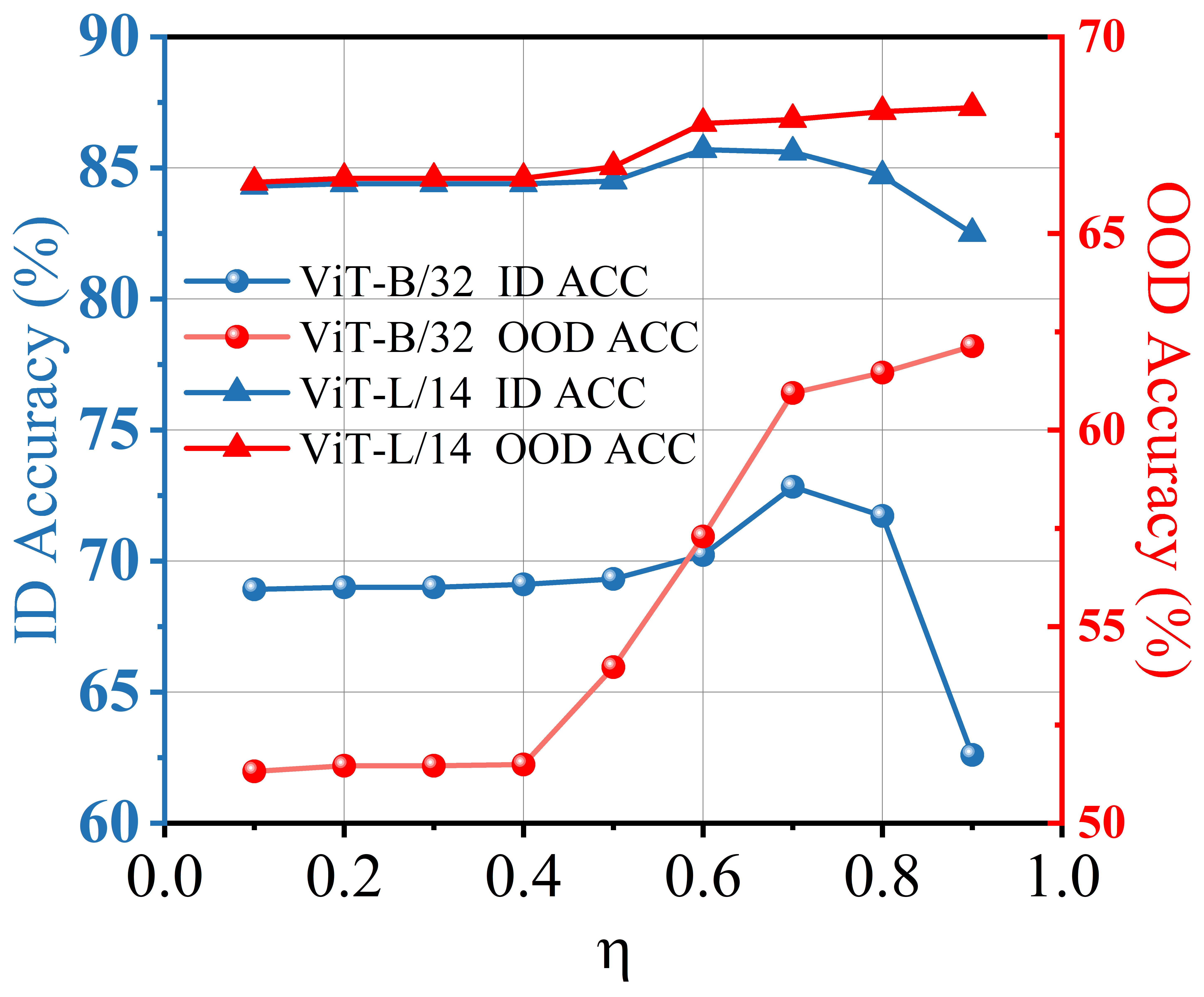}
     \vspace{-1.5em} 
\caption{\small{Performance of Task Arithmetic+\method with varying $\eta$.}}
    \label{fig:eta}
  \end{minipage}
  \vspace{-20pt}
\end{figure}



\vspace{1em}

\textbf{Effect of our components.} We conduct ablation studies based on Task Arithmetic and LW AdaMerging with ViT-B/32. \method includes three components: salience score, OR operation ($\lor$), and coefficient scaling (scale) for LW AdaMerging, which learns the layer-wise merging weights. To systematically investigate the contribution of each individual component, we  remove each component and observe the resultant impact. 
 Especially, for w/o saliency score, we design the pruning metric based on the mean absolute value of the task vectors across each layer i.e., $\mathbb{E}(|\boldsymbol{\tau}^l_k|) $. The results are illustrated in Fig.\ref{fig:ablation_study}; see details in Tab. \ref{tab:ablation_ID} and Tab. \ref{tab:ablation_OOD} of Appendix. We can find that w/o saliency score but using $\mathbb{E}(|\boldsymbol{\tau}^l_k|) $  can degrade the H-score a lot, where the performance drop is mainly on ID. Besides, those variants with saliency score are better than original merging model methods, indicating the effectiveness of our designed salience score in balancing ID and OOD. In addition, w/o OR operation means that we discard more layer-wise parameters in task vectors depending only on saliency score, which also reduces the H-score of merged model.  When combining ours with LW AdaMerging, removing scaling srategy is also harmful for the merged model. It proves the effectiveness of our proposed OR operation and scaling strategy.

\textbf{Comparison with OOD methods for fine-tuned models.} To explore the advantages of ours, we  consider following methods specifically designed for enhancing the OOD performance of fine-tuned models: WiSE-FT \cite{wortsman2022robust}, Model Stock \cite{jang2024model}, and LiNeS \cite{wang2024lines}. Among them, WiSE-FT and Model Stock are specifically designed for the CLIP architecture, and LiNeS targets the ViT architecture within CLIP. We implement these methods on fine-tuned models and then merge the multiple fine-tuned models with Task Arithmetic. As listed in Table \ref{table:OOD and Pruning}, all methods can enhance both ID and OOD performance of the merged model, but ours performs best. As a strong baseline, LiNeS outperforms \method in terms of ID performance, which  can be attributed to its amplification  of task-specific parameters. However, it is still inferior to \method about the OOD performance and overall H-score.  Additionally, for different methods and architectures, LiNes requires the  validation set to determine the optimal scaling coefficient, whereas our pruning ratio of 0.7 is generally applicable. This further demonstrates the advantage of \method in balancing the ID and OOD performance of the merged model.  See more details in Appendix \ref{sec:OOD exc}.

\textbf{Comparison with other pruning methods.} To investigate the superiority of our method over other pruning methods, we employ the following pruning baselines: (1) DARE \cite{yu2024language}, Drops delta parameters with a ratio p And REscales; (2) Magnitude-based Weight Pruning (MWP) \cite{sanh2020movement}, which determines the mask based on the absolute value of each parameter within task vector; (3) random mask, wherein the mask is stochastically generated via a Bernoulli distribution; (4) the mean absolute value of the task vectors per layer, denoted as absolute value. As depicted in Tab. \ref{table:OOD and Pruning}, \method's H-score exceeds that of the most favorable Magnitude-based Weight Pruning among the four baselines by 3.2\%. Although random mask exhibits superior performance on the OOD dataset, it incurs a significant performance deterioration on the ID dataset. This demonstrates that \method's
effectiveness of proposed salience score in pruning parameters. See more details in Appendix \ref{sec:Pruning}.

\begin{wraptable}[14]{r}{6cm}
\vspace{-1.5em}
\centering
\caption{ \small{Comparison with other OOD methods (above) and purning methods (below) on  ViT-B/32.}}
\label{table:OOD and Pruning}
\scriptsize
\setlength{\tabcolsep}{2pt}
\scalebox{0.90}{
\begin{tabular}{@{}c|cccc@{}}
\toprule
 \multicolumn{2}{c}{\textbf{Performace (\%) (→)}}  &  \multirow{2}{*}{\textbf{ID. Avg}} & \multirow{2}{*}{\textbf{OOD. Avg}}&\multirow{2}{*}{\textbf{H-score}} \\
\multicolumn{2}{c}{\textbf{Method (↓)}} \\ 
\midrule
Baseline&Task Arithmetic \cite{ilharco2022editing} 
& 69.1&  51.3& 58.9   \\  
\midrule
\multicolumn{1}{>{\cellcolor{lightgray!10}}c|}{\multirow{3}{*}{OOD}} & \cellcolor{lightgray!10}w/ WiSE-FT \cite{wortsman2022robust}
\cellcolor{lightgray!10}& \cellcolor{lightgray!10}70.5 & \cellcolor{lightgray!10}55.4 & \cellcolor{lightgray!10}62.0   \\ 

\multicolumn{1}{>{\cellcolor{lightgray!10}}c|}{OOD}& \cellcolor{lightgray!10}w/ Model Stock \cite{jang2024model}
& \cellcolor{lightgray!10}72.0 &\cellcolor{lightgray!10}58.7 & \cellcolor{lightgray!10}64.7   \\ 
\multicolumn{1}{>{\cellcolor{lightgray!10}}c|}{} & \cellcolor{lightgray!10}w/ LiNeS  \cite{wang2024lines}
& \cellcolor{lightgray!10}74.1& \cellcolor{lightgray!10}58.0 & \cellcolor{lightgray!10}65.1   \\

\midrule


 \rowcolor{lightgray!10}& w/ DARE \cite{yu2024language}
& 49.1 &  61.5 &  54.6  \\ 
\rowcolor{lightgray!10}& w/ MWP \cite{sanh2020movement}
& 71.7 &  56.4 &63.1   \\ 
\rowcolor{lightgray!10} & w/ random mask
& 49.1 & 61.7 & 54.7  \\ 
\rowcolor{lightgray!10} \multirow{-4}{*}{Pruning} & w/ absolute  value 
& 48.9 & 61.5 & 54.5  \\

\midrule
\textbf{Ours}&\textbf{ w/ LwPT} & 72.8 &60.9  & \textbf{66.3} \textsuperscript{\best{(+7.4)}}  \\ 
\bottomrule
\end{tabular}
}
\end{wraptable}

\vspace{-2mm}
\subsection{Additional analysis}
\vspace{-2mm}

\textbf{Trade-off between ID and OOD.} To explore the impact of pruning ratio $\eta$ on the trade-off between ID and OOD task performance, we conducted the experiments using {Task Arithmetic w/ \method} on the ViT-B/32 and ViT-L/14. Notably, $\eta$ serves as a parameter to balance the  pretrained model and task vectors of the fine-tuned models.  $\eta=1$  signifies a predominance of the pretrained model, while $\eta=0$  represents the original merging model. We can see that the OOD performance gradually increases with a development pruning ratio. 
Conversely, the ID task performance shows an initial improvement followed by a decline. It is reasonable since a proper pruning ratio can increase the ID and OOD performance for pruning redundant parameters and improve the generalization performance. And a too large pruning ratio will prune the useful parameters for ID tasks and hurt its performance, indicating the trade-off between ID and OOD tasks. Besides, a pruning ratio $\eta$ between 0.6 and 0.8 offers a relatively desired trade-off, ensuring improved OOD performance while maintaining desired ID accuracy, where we set $\eta=0.7$ for all experiments.

\begin{figure}[htbp!]
  \centering
  \begin{minipage}[t]{0.68\textwidth}
    \centering
    \includegraphics[width=\textwidth]{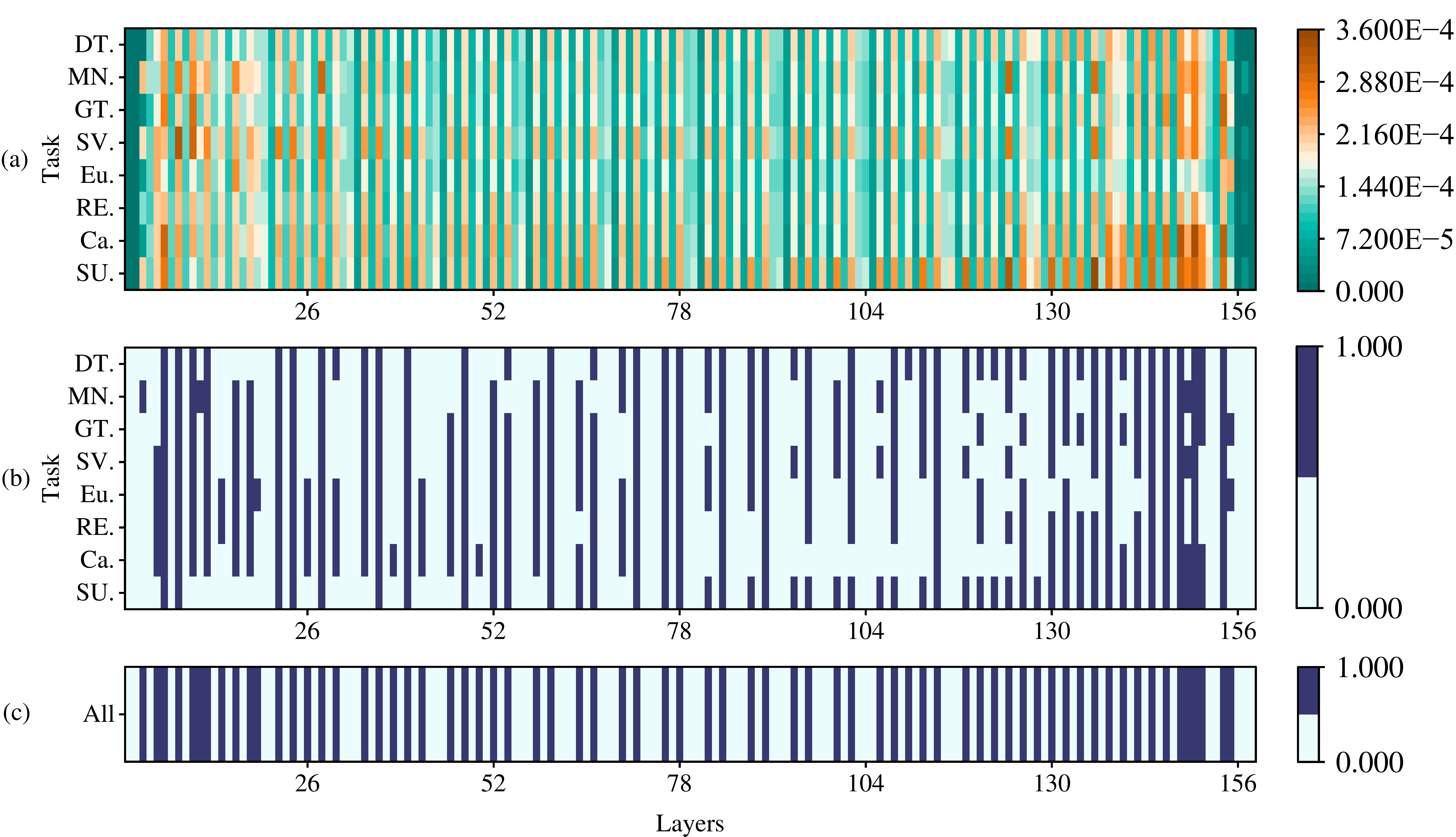}  
    \vspace{-8pt}
    \caption{\small{Visualizing of (a) salience score matrix, (b) mask vector for each task, and (c) final mask vector, all on VIT-B/32, where x-axis denotes the layer index, y-axis in (a-b) denotes task name.}}
        \vspace{-10pt}
    \label{Significance score Vit-B}
  \end{minipage}
   \hfill  
  \begin{minipage}[t]{0.30\textwidth}
    \centering
    \includegraphics[width=\linewidth]{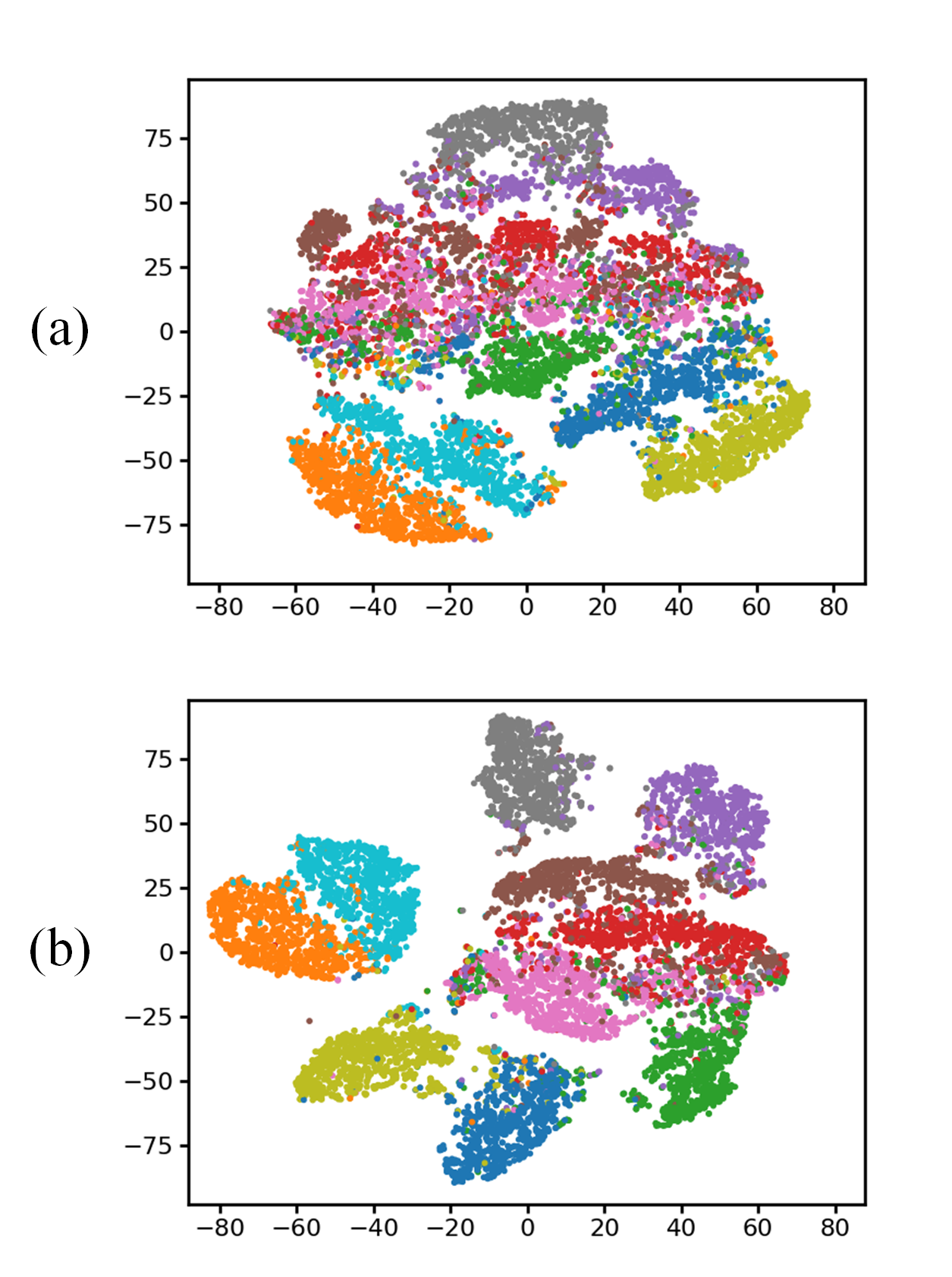}
     \vspace{-10pt}
    \caption{\small{T-SNE visualizations of (a) Task Arithmetic, and (b) Task Arithmetic w/ \method on CIFAR10.}}
\label{fig:T-sne_cifar10}
  \end{minipage}
  \hfill
  \vspace{-15pt}
\end{figure}

\textbf{Saliency score and mask vectors.} To facilitate an intuitive analysis of saliency score, masks associated with each task vector and their interrelationships, we visualize them for ViT-B/32 in Fig.\ref{Significance score Vit-B}. We can see that, for different tasks, the salience score vectors usually have low values in certain same layers, indicating their redundancy in these layers. Besides, in each salience score vector, most of the elements have low values, which explains why the model produces better OOD and ID performance even when we throw away most layers (pruning ratio $\eta=0.7$). Therefore, the mask vectors of different tasks have a lot of overlapping layers, which can be viewed as those parameters unimportant to all tasks. Those non-overlapping mask information might be viewed as the beneficial parameters for the corresponding ID task, which is the reason that we choose to keep them. Therefore, we adopt the OR operation to extract a shared mask vector from all masks, i.e., replacing these overlapping layers with the corresponding layers of pre-trained model.
 To explore wether ours can extract the discriminative features intuitively, we present a T-SNE visualization for features from Task Arithmetic and the Task Arithmetic+\method on OOD tasks  in Fig.\ref{fig:T-sne_cifar10}. 
Compared to Task Arithmetic, introducing our  \method can enhance the separability of representations   across different categories. Due to the limited space, we provide additional results in Appendix \ref{ExpDel} and Appendix \ref{Visualization Results}, such as storage overhead, etc.





\vspace{-4mm}
\section{Conclusion} \label{conclusion}
\vspace{-2mm}

In this work, we first systematically analyze existing model merging techniques and identify their limitations in generalizing to OOD data. To address this issue, we propose a novel plug-and-play framework, termed LwPTV, and introduce a training-free saliency score to measure the redundancy of task vectors. By pruning the layer-wise parameters with low saliency score, ours enhances the OOD generalization capability of merged models while preserving ID performance. Extensive experimental results demonstrate that our approach significantly improves the OOD generalization performance of existing  model merging algorithms.

\section*{Impact Statement}
This paper presents work whose goal is to advance the field of 
model merging in machine learning. There are many potential societal consequences 
of our work, none which we feel must be specifically highlighted here.

\newpage

\appendix

\section{Theoretical analyses}
\subsection{Proof of proposition 1}
\label{sec:Proposition 1}

We start by introducing the theoretical setting. 

\textbf{Theoretical formulation. } 
Consider a binary classification problem, where the target is to predict $y\in\{+1,-1\}$ given the input $\boldsymbol{X}=(\boldsymbol{x}_1,\boldsymbol{x}_2,\cdots,\boldsymbol{x}_P)\in\mathbb{R}^{d\times P}$ that contains $P$ $d$-dimensional tokens. Following the state-of-the-art theoretical works \citep{LWLC23, MBGN24, LWLC24, JHZS24} on the generalization and learning dynamics of neural models, we use a one-layer single-head Transformer as the learner model for theoretical analysis, which can be characterized as $f(\boldsymbol{X};\boldsymbol{\theta})=1/P\sum_{l=1}^P \boldsymbol{a}_{l}^\top\text{Relu}(\boldsymbol{V}\sum_{s=1}^P\boldsymbol{x}_s\text{softmax}_l(\boldsymbol{x}_s^\top\boldsymbol{W}\boldsymbol{x}_l))$, where $\boldsymbol{\theta}=\{\{\boldsymbol{a}_l\}_{l=1}^P, \boldsymbol{V}, \boldsymbol{W}\}$ with $\boldsymbol{a}_l\in\mathbb{R}^{m}$, $\boldsymbol{V}\in\mathbb{R}^{m\times d}$, and $\boldsymbol{W}\in\mathbb{R}^{d\times d}$. $\text{softmax}_l(\boldsymbol{x}_s^\top\boldsymbol{W}\boldsymbol{x}_l)=e^{\boldsymbol{x}_s^\top\boldsymbol{W}\boldsymbol{x}_l}/\sum_{j=1}^P e^{\boldsymbol{x}_j^\top\boldsymbol{W}\boldsymbol{x}_l}$. $\text{Relu}(\boldsymbol{z})=\max\{0,\boldsymbol{z}\}$.

Denote $\mathcal{T}_k$, $k\leq K$ as the $i$-th task function that we aim to learn with the training dataset $\{\boldsymbol{X}^n, y^n\}_{n=1}^N$. Starting from the initialization $\boldsymbol{\theta}_{pre}$, we train the model using stochastic gradient descent with Hinge loss to obtain the fine-tuned model $\boldsymbol{\theta}_i$. The task vector is then computed as $\boldsymbol{\tau}_k=\boldsymbol{\theta}_k-\boldsymbol{\theta}_{pre}$. 

The data formulation follows Definition 2 of \citep{LZZC25}, where the label of each data for task $\mathcal{T}_k$ is determined by the majority between tokens with two discriminative features $\boldsymbol{\mu}_{\mathcal{T}_k}$ and $-\boldsymbol{\mu}_{\mathcal{T}_k}$. If the number of tokens equal to $\boldsymbol{\mu}_{\mathcal{T}_k}$ (or $-\boldsymbol{\mu}_{\mathcal{T}_k}$) is large than that of $-\boldsymbol{\mu}_{\mathcal{T}_k}$ (or $\boldsymbol{\mu}_{\mathcal{T}_k}$), then $y=+1$ (or $y=-1$). Each data can also contain task-irrelevant tokens from an orthonormal set $\{\boldsymbol{v}_1,\boldsymbol{v}_2,\cdots,\boldsymbol{v}_M\}$ that are orthogonal to discriminative features. Given $\boldsymbol{\tau}_k$, $k\in [K]$, trained with a batch size of $B\geq \Omega(\epsilon^{-2}\log M)$ to achieve an $\epsilon$ generalization error on $\mathcal{T}_k$, denote $\boldsymbol{u}_k^i$ as the $i$-th row of the part of $\boldsymbol{V}$ in $\boldsymbol{\tau}_k$. Then, we can prove Proposition \ref{prop: condition} as follows.

\begin{proof}

By Corollary 2 and Lemma 5 of \citep{LZZC25}, we know that with a constant probability, which we denote by $p$, $\boldsymbol{u}_k^i$, $k\in[K]$, $i\in[m]$, leans the discriminative feature. Then, with a high probability of $1-p^K$, there exists $k_1\neq k_2\in[K]$, such that the $i$-th neuron of $\boldsymbol{\tau}_{k_1}$ learns discriminative features, while the $i$-th neuron of $\boldsymbol{\tau}_{k_2}$ fail to learn discriminative features. 

By their Definition 4 of \citep{LZZC25}, we have that for any $i\mathcal{S}$,
\begin{equation}
    \|\boldsymbol{u}_k^i\|\geq \Omega(m^{-1/2}). \label{lucky}
\end{equation}

By Lemma 5 in \citep{LZZC25}, we know that neurons of $\boldsymbol{\tau}_{k_1}$ in $\mathcal{S}$ are mainly in the direction of $\boldsymbol{\mu}_{\mathcal{T}_k}$ or $-\boldsymbol{\mu}_{\mathcal{T}_k}$, with the norm of all other directions smaller than a $1/\sqrt{M}$ scaling of that in the direction of the discriminative pattern. Then, the variance of lucky neurons in different task vectors can be computed as
\begin{equation}
\begin{aligned}
    \text{DV}(g(\boldsymbol{\tau}_{k_1}, \mathcal{S});\{\boldsymbol{\tau}_j\}_{j=1}^K)=& \mathbb{E}  
\Big[  \Big\| g(\boldsymbol{\tau}_{k_1}, \mathcal{S}) - \frac{1}{K} \sum_{j=1}^K g(\boldsymbol{\tau}_j, \mathcal{S}) \Big\|  \Big]\\
    \geq & |\mathcal{S}|\cdot \Omega(m^{-1}).\label{lucky var}
\end{aligned}
\end{equation}
Hence, the diversity of neurons that learn discriminative patterns is in the order of $|\mathcal{S}|\cdot \Omega(m^{-1/2})$.

For neurons that learn no discriminative features, $k\in[K]$, $i\in[m]$, we have
\begin{equation}
\begin{aligned}
    \text{DV}(g(\boldsymbol{\tau}_{k_2}, \mathcal{S});\{\boldsymbol{\tau}_j\}_{j=1}^K)=& \mathbb{E}  
\Big[  \Big\| g(\boldsymbol{\tau}_{k_2}, \mathcal{S}) - \frac{1}{K} \sum_{j=1}^K g(\boldsymbol{\tau}_j, \mathcal{S}) \Big\|  \Big]\\
    \leq & |\mathcal{S}|\cdot O(m^{-1})\cdot \frac{1}{\sqrt{K}}.\label{unlucky var1}
\end{aligned}
\end{equation}
The last step comes from that (i) $\|g(\boldsymbol{\tau}_{k_2}, \mathcal{S})\|\leq O(m^{-1/2})\sqrt{\log B/B}$. (ii) 

\begin{equation}
    \Big\|\frac{1}{K} \sum_{j=1}^K g(\boldsymbol{\tau}_j, \mathcal{S})\Big\|\leq O(m^{-1/2})\cdot \frac{1}{\sqrt{K}}.\label{unlucky var}
\end{equation}
Then, by combining (\ref{lucky var}) and (\ref{unlucky var}), we have
\begin{equation}
    \text{DV}(g(\boldsymbol{\tau}_{k_1}, \mathcal{S});\{\boldsymbol{\tau}_j\}_{j=1}^K)\geq \sqrt{K}\cdot \Omega(\text{DV}(g(\boldsymbol{\tau}_{k_2}, \mathcal{S});\{\boldsymbol{\tau}_j\}_{j=1}^K)).
\end{equation}

\end{proof}

\section{Experimental details}\label{ExpDel}

\subsection{The analysis of scaling}
\label{sub:analysis_eta}

\begin{figure}[H]
  \centering

  \begin{subfigure}[b]{0.45\textwidth}
    \includegraphics[width=\linewidth]{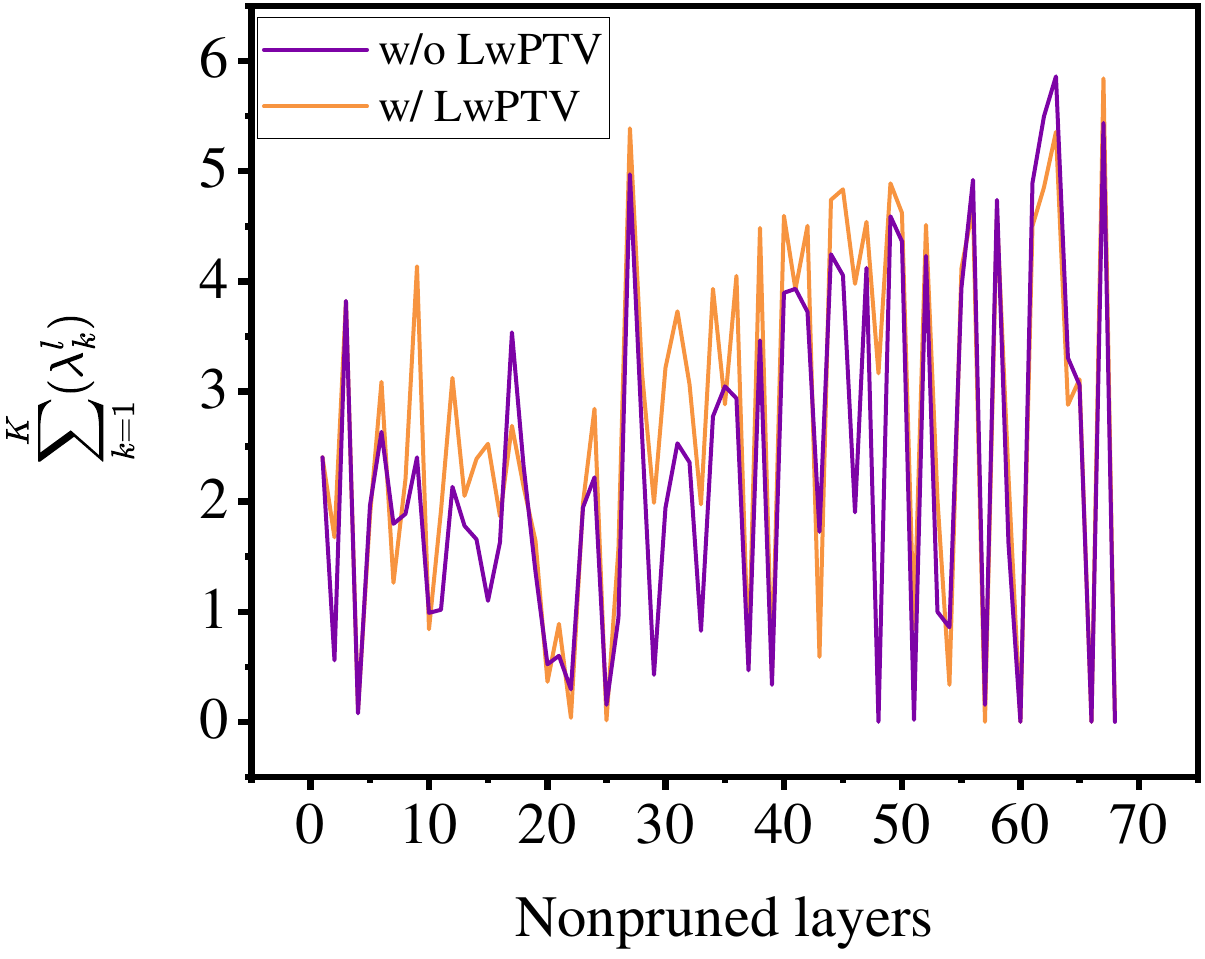}
    \caption{Lw AdaMerging w/ \method \textit{vs.} Lw AdaMerging on ViT-B/32.} 
    \label{fig:lambda_sum_ViT_B_32}
  \end{subfigure}
   \hfill
  \begin{subfigure}[b]{0.45\textwidth}
    \includegraphics[width=\linewidth]{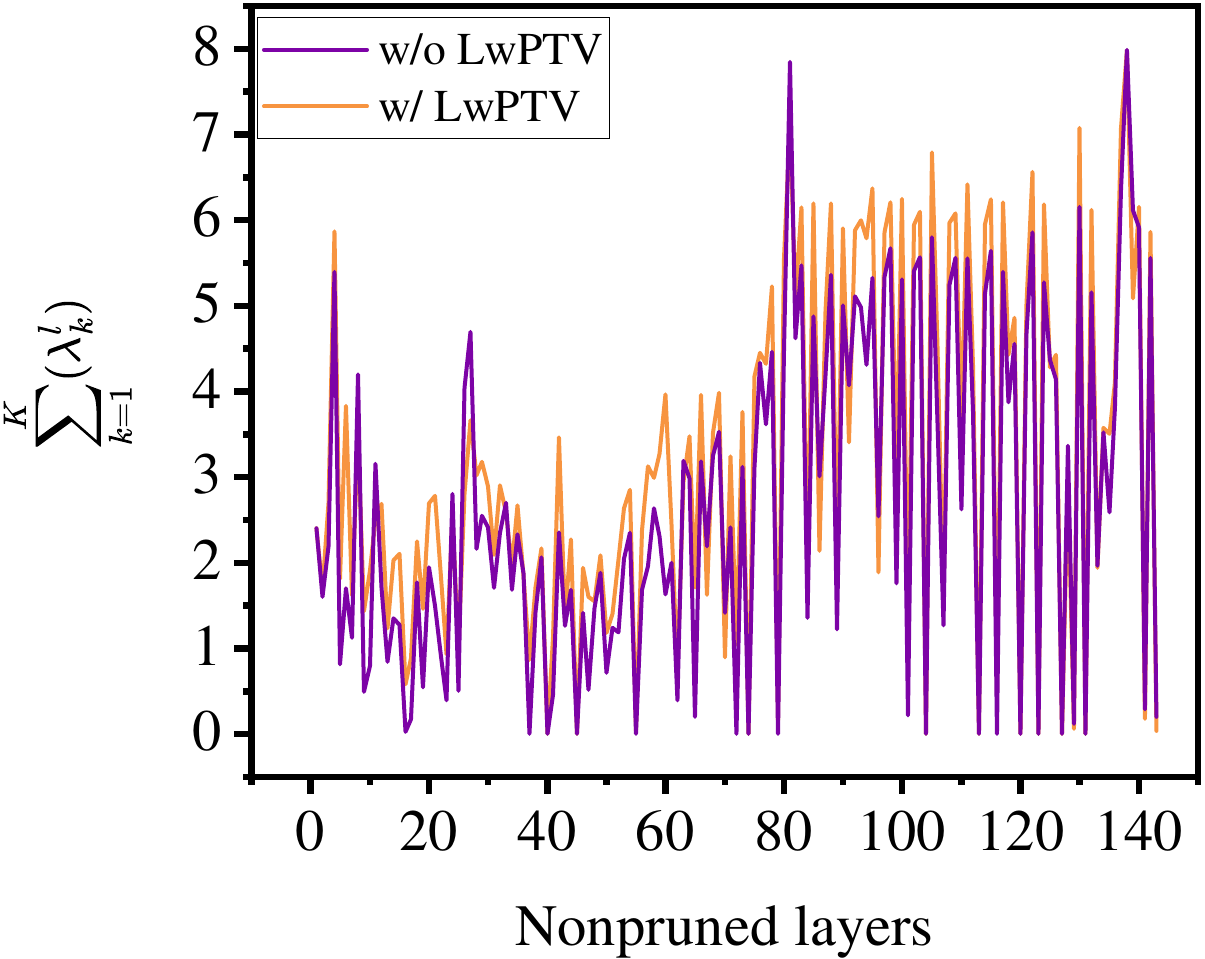}
    \caption{Lw AdaMerging w/ \method \textit{vs.} Lw AdaMerging on ViT-L/14.} 
    \label{fig:lambda_sum_ViT_L_14}
  \end{subfigure}
  \hfill



   \caption{The variations in $\lambda^l_k$
  of unpruned layers subsequent to the application of the LW AdaMerging and LW AdaMerging w/ LwPTV.}

  \label{fig:lambda_variation}
\end{figure}

To investigate the impact of \method (w/o scale) on the layer-wise coefficients $\lambda^l_k$ of LW AdaMerging, we have plotted a comparative figure of the unpruned layer coefficients under two scenarios: LW AdaMerging and LW AdaMerging w/ LwPTV. The results are illustrated in Fig.\ref{fig:lambda_sum_ViT_B_32} and Fig.\ref{fig:lambda_sum_ViT_L_14}. In these figures, $\sum^K_{k=1}(\lambda^l_k)$
  represents the sum of the coefficients of $K$ tasks on the $l$-th layer. 
It can be clearly observed from these two figures that the $\lambda^l_k$ obtained after pruning and optimization are significantly increased. This is because, during the process of optimizing $\lambda^l_k$ by entropy minimization, the adjustable coefficients $\lambda^l_k$ are amplified to enhance the influence of unpruned parameters in order to minimize the loss function as much as possible. This allows the merged model to maintain its ID performance even as the pruning ratio continuously increases, but its OOD performance does not show a significant improvement.
In order to improve the OOD performance, we further use   $\hat{\lambda}^l_k=\eta \cdot \lambda^l_k$ to scale the coefficient after optimizing the \eqref{Eq3-5}.

\subsection{Algorithm}
\label{sub:algorithmic2}
For Task Arithmetic, Ties, TW AdaMerging, TW AdaMerging++, LW AdaMerging, LW AdaMerging++, and PCB-MERGING, we present the algorithmic workflows of their integration with \method in Algorithm \ref{alg2}.

\begin{algorithm}[htbp]
   \caption{Overall algorithm when combining our \method with existing model merging methods.}
   \label{alg2}
\begin{algorithmic}[1]
   \State {\bfseries Input:} pre-trained model parameters $\boldsymbol{\theta}_{pre}$, $K$ task-specific fine-tuned model parameters $ \{\boldsymbol{\theta}_{k}\}^K_{k=1}$ and hyperparameter $\eta$;
   \For{$k=1$ {\bfseries to} $K$}
   \State Compute the task vector $\boldsymbol{\tau}_k$ : \State   \:\:\:\: $\boldsymbol{\tau}_k=\boldsymbol{\theta}_k-\boldsymbol{\theta}_{pre}$;
   \State Compute the salience score $\mathbf{s}_k$ of $\boldsymbol{\tau}_k$ by \eqref{Eq3-3};
   \State Compute the mask $\mathbf{m}_k$ by \eqref{Eq3-4};
   \EndFor
   \State Calculate the final mask $\hat{\mathbf{m}}$ by \eqref{Eq3-6}.
   \\
   \# Model merging methods execute:
    \State \textbf{If Task Arithmetic}: $\hat{\boldsymbol{\theta}}_m=\boldsymbol{\theta}_{pre}+\hat{\mathbf{m}} \odot \lambda \sum_{k=1}^K \boldsymbol{\tau}_k$.
    \State \textbf{If Ties}: 
     Perform trim and sign election operations on $\{\hat{\mathbf{m}}\odot\boldsymbol{\tau}_k\}_{k=1}^K$  to obtain $\{\hat{\boldsymbol{\tau}}_k\}^K_{k=1}$, then perform $\hat{\boldsymbol{\theta}}_m=\boldsymbol{\theta}_{pre}+\lambda \sum_{k=1}^K \hat{\boldsymbol{\tau}}_k$.

      \State \textbf{If TW AdaMerging}: Get $\{ \lambda_k\}^K_{k=1}$ by minimizing  entropy base on \eqref{Eq3-5} , then perform \eqref{Eq3-5}. 
       \State \textbf{If TW AdaMerging++}: Perform trim and sign election operations on $\{\hat{\mathbf{m}}\odot\boldsymbol{\tau}_k\}_{k=1}^K$ to obtain $\{\hat{\boldsymbol{\tau}}_k\}^K_{k=1}$, then get $\{ \lambda_k\}^K_{k=1}$ by minimizing  entropy base on $\boldsymbol{\theta}_{pre}+ \sum_{k=1}^K \lambda_k \hat{\boldsymbol{\tau}}_k$ , then perform  $\hat{\boldsymbol{\theta}}_m=\boldsymbol{\theta}_{pre}+ \sum_{k=1}^K \lambda_k \hat{\boldsymbol{\tau}}_k$.

       \State \textbf{If LW AdaMerging}: Get $\{\{\lambda_k^l\}^L_{l=1}\}^K_{k=1}$ by minimizing  entropy base on \eqref{Eq3-5}, then perform  $\hat{\boldsymbol{\theta}}_m=\boldsymbol{\theta}_{pre}+\hat{\mathbf{m}} \odot \{\sum_{k=1}^K \eta\lambda^l_k \boldsymbol{\tau}_k\}^L_{l=1}$. 
       \State \textbf{If LW AdaMerging++}:   Perform trim and sign election operations on $\{\hat{\mathbf{m}}\odot\boldsymbol{\tau}_k\}_{k=1}^K$ to obtain $\{\hat{\boldsymbol{\tau}}_k\}^K_{k=1}$, and then get $\{\{\lambda_k^l\}^L_{l=1}\}^K_{k=1}$ by minimizing  entropy base on $\boldsymbol{\theta}_{pre}+ \{\sum_{k=1}^K \lambda^l_k \hat{\boldsymbol{\tau}}_k\}^L_{l=1}$, then perform  $\hat{\boldsymbol{\theta}}_m=\boldsymbol{\theta}_{pre}+ \{\sum_{k=1}^K \eta\lambda^l_k \hat{\boldsymbol{\tau}}_k\}^L_{l=1}$. 
        \State \textbf{If PCB}: Base on $\{\hat{\mathbf{m}}\odot\boldsymbol{\tau}_k\}_{k=1}^K$ to get importance score $\{\hat{\boldsymbol{\beta}}_k\}^K_{k=1}$, then perform $\hat{\boldsymbol{\theta}}_m=\boldsymbol{\theta}_{pre}+\hat{\mathbf{m}} \odot \sum_{k=1}^K (\hat{\boldsymbol{\beta}}_k \odot\lambda_k \boldsymbol{\tau}_k)/\sum_{k=1}^K\hat{\boldsymbol{\beta}}_i$.

\end{algorithmic}

\end{algorithm}

\subsection{Dataset details}\label{sub:datasets}  Following Yadav et al. \cite{yadav2023resolving, zhou2022learning}, we utilize eight datasets,  including Stanford Cars \cite{krause20133d}, DTD \cite{cimpoi2014describing}, EuroSAT \cite{helber2019eurosat}, GTSRB \cite{stallkamp2011german}, MNIST \cite{deng2012mnist}, RESISC45 \cite{cheng2017remote}, SUN397 \cite{xiao2016sun}, and SVHN \cite{netzer2011reading},  for merging vision models. These datasets are considered ID datasets. Following Wang et al. \cite{wanglocalizing}, we select  CIFAR100 \cite{krizhevsky2009learning}, STL10 \cite{coates2011analysis}, Flowers102 \cite{nilsback2008automated}, OxfordIIITPet \cite{parkhi2012cats}, PCAM \cite{veeling2018rotation}, FER2013 \cite{goodfellow2013challenges}, EMNIST \cite{cohen2017emnist}, CIFAR10 \cite{krizhevsky2009learning}, Food101 \cite{bossard2014food}, FashionMNIST \cite{xiao2017fashion}, RenderedSST2 \cite{socher2013recursive, radford2019language}, KMNIST \cite{clanuwat2018deep} and ImageNet-1k \cite{deng2009imagenet} as OOD datasets. The details of these datasets are shown in Tab.\ref{In Domain and Out of Domain Datasets details}

\begin{table}[h]
\centering
\caption{In Domain and Out of Domain datasets details}
\label{In Domain and Out of Domain Datasets details}
\setlength{\tabcolsep}{1mm}
\resizebox{\textwidth}{!}{
\begin{tabular}{ccccccc}
\toprule
\rowcolor{gray!30} \multirow{1}{*}{\textbf{Feature(→)}} & \multicolumn{6}{l}{} 
 \\
 \rowcolor{gray!30} \multirow{1}{*}{\textbf{Dataset (↓)}} & \multirow{-2}{*}{\textbf{Type}}& \multirow{-2}{*}{\textbf{Sample Size}} & \multirow{-2}{*}{\textbf{Test Set Sample Size}} & \multirow{-2}{*}{\textbf{Number of Classes}} & \multirow{-2}{*}{\textbf{Samples per Class}} & \multirow{-2}{*}{\textbf{Class Balance}}\\
\midrule
\multirow{1}{*}{\textbf{Stanford Cars}} & \multirow{8}{*}{\textbf{ID}} & 16185 & 8041 & 196 & 1:1 split & Yes \\
\textbf{DTD} &  & 5640 & 1880 & 47 & 40  & Yes \\
\textbf{EuroSAT} & & 27000 & 2700 & 10 & 2000-3000 & Yes\\
\textbf{GTSRB} & & 51839 & 12630 & 43 & / & No\\
\textbf{MNIST} & & 70000 & 10000 & 10 & 1000 & Yes\\
\textbf{RESISC45} & & 31500 & 6300 & 45 & approximately 700 & Yes \\
\textbf{SUN397} & & 108754 & 7940 & 397 & at least 100 images per category & No\\
\textbf{SVHN} & & 60000 & 26032 & 10 & / & No\\
\midrule
\textbf{CIFAR100} & \multirow{12}{*}{\textbf{OOD}}& 60000 & 10000 & 100 & 100 &Yes\\
\textbf{STL10} & & 13000 & 8000 & 10 & 800 & Yes\\
\textbf{Flowers102} & & 8189 & 6129 & 102 & per class 40-250 & No\\
\textbf{OxfordIIITPet} && 7349 & 3669 & 37 & approximately 100 & Yes\\
\textbf{PCAM} && 327680 & 40960 & 2  & / &/ \\
\textbf{FER2013} && 35887 & 7178 & 7 &/ & No\\
\textbf{EMNIST} && 280000 & 40000 & 10 & 4000 & Yes\\
\textbf{CIFAR10} && 60000 & 10000 & 10 & 1000 & Yes\\
\textbf{Food101} && 101000 & 25250 & 101 & 250 & Yes\\
\textbf{FashionMNIST} && 70000 & 10000 & 10 & 1000 & Yes\\
\textbf{RenderedSST2} && 9613 & 1821 & 2 & / & /\\
\textbf{KMNIST} && 70000 & 10000 & 10 & 1000 & Yes\\
\textbf{ImageNet1000} && 3200000 & 50000 & 1000 & / & No\\
\bottomrule
\end{tabular}
}
\end{table}

\subsection{Evaluation metrics}\label{sub:Evaluation_Metrics}

To rigorously assess the efficacy of our method in enhancing the generalization capability of model merging methods on OOD datasets, while concurrently preserving their efficacy on ID datasets, we utilize a key evaluation metrics: H-score. The H-score is defined by the harmonic mean of the mean performance for the ID datasets, denoted as $Avg(P_{ID})$,  coupled with the mean performance for the OOD datasets, denoted as $Avg(P_{OOD})$. It can be formulated as follows:

\begin{eqnarray}
P_H=\frac{2 \times Avg(P_{ID}) \times Avg(P_{OOD})}{Avg(P_{ID}) + Avg(P_{OOD})}.
\label{Eq6-1}
\end{eqnarray}

\subsection{ID performance}
\label{sec:ID_Per}
We evaluate the performance of our proposed method alongside baseline approaches on ID datasets across three distinct architectures, namely ViT-B/32, ViT-L/14, and ViT-H/14. The corresponding results are systematically summarized in Tab.\ref{ViT-B-32-ID} and Tab.\ref{ViT-H-14-ID}, respectively. 
The experimental results demonstrate that our method, when integrated with existing model merging methods, induces minimal performance degradation on ID datasets, and in some cases even enhances ID performance. This improvement can be attributed to the reduced parameter conflicts achieved through our task vector pruning strategy, which effectively mitigates interference between different task-specific parameters.

\begin{table*}[htbp]
\centering
\caption{Performance comparison for ID datasets on ViT-B/32 and ViT-L/14.}
\label{ViT-B-32-ID}
\small
\setlength{\tabcolsep}{0.5mm}
\resizebox{\textwidth}{!}{
\begin{tabular}{ccccccccccc}
\toprule
\multicolumn{2}{c}{\textbf{Method (↓)}} & \textbf{SUN397} & \textbf{Cars} & \textbf{RESISC45} & \textbf{EuroSAT} & \textbf{SVHN} & \textbf{GTSRB}& \textbf{MNIST}& \textbf{DTD}& \textbf{Avg.} \\ 
\midrule 
\multirow{20}{*}{\raisebox{-15mm}{\rotatebox{90}{\textbf{ViT-B/32}}}}
&
Pretrained & 62.3 & 59.7 & 60.7 & 45.5 &31.4 &32.6& 48.5 &43.8 &48.0 \\ 

&Individual  & 75.3 & 77.7 & 96.1 & 99.7 & 97.5 & 98.7& 99.7& 79.4 &90.5 \\ 


\cmidrule(lr){2-11} 
&Weight Averaging \cite{wortsman2022model} & 65.3 & 63.4 & 71.4 & 71.7 & 64.2& 52.8& 87.5 &50.1& 65.8 \\ 

&Fisher Merging \cite{matena2022merging}  & 68.6 & 69.2& 70.7& 66.4& 72.9& 51.1& 87.9& 59.9 &68.3\\

&RegMean \cite{jindataless} & 65.3 & 63.5& 75.6& 78.6& 78.1 &67.4 &93.7& 52.0& 71.8 \\ 
&AdaMerging w/ Surgery \cite{yangrepresentation} &69.8& 71.0 &88.9& 98.1& 91.7& 96.5& 98.8 &73.6& 86.1 \\ 
\cmidrule(lr){2-11} 

&Task Arithmetic \cite{ilharco2022editing} & 55.2& 54.9& 66.7& 78.9& \textbf{80.2}& \textbf{69.7}& \textbf{97.3}& 50.4 &69.1 \\ 
&\cellcolor{gray!10} \textbf{w/ \method} (Ours) &\cellcolor{gray!10}\textbf{66.8} &\cellcolor{gray!10}\textbf{66.0} &\cellcolor{gray!10}\textbf{77.2}& \cellcolor{gray!10}\textbf{79.1}&\cellcolor{gray!10}78.3& \cellcolor{gray!10}68.7& \cellcolor{gray!10}92.3& \cellcolor{gray!10}\textbf{54.4}& \cellcolor{gray!10}\textbf{72.8}\textsuperscript{\best{(+3.7)}} \\ 
\cmidrule(lr){2-11}

 &Ties \cite{yadav2023resolving} & 65.0 & 64.4 &74.8 &77.4& 
 \textbf{81.2}& \textbf{69.3}& \textbf{96.5}& 54.5 &\textbf{72.9}\\ 
 &\cellcolor{gray!10} \textbf{w/ \method} (Ours) & \cellcolor{gray!10}\textbf{67.6} & \cellcolor{gray!10}\textbf{66.7 }&\cellcolor{gray!10}\textbf{77.4} &\cellcolor{gray!10}\textbf{77.4}& \cellcolor{gray!10}77.7&\cellcolor{gray!10}66.8&\cellcolor{gray!10}92.4&\cellcolor{gray!10}\textbf{55.3 }&\cellcolor{gray!10}72.7\textsuperscript{\color[RGB]{255,0,0}{(-0.2)}}\\ 
\cmidrule(lr){2-11} 

&TW AdaMerging \cite{yangadamerging} &58.0 &53.2 &68.8 &85.7& 81.1 &84.4& 92.4& 44.8& 71.1\\ 
&\cellcolor{gray!10}\textbf{w/ \method} (Ours) &\cellcolor{gray!10}\textbf{61.6} &\cellcolor{gray!10}\textbf{59.7}&\cellcolor{gray!10}\textbf{77.0} &\cellcolor{gray!10}\textbf{87.3}&\cellcolor{gray!10}\textbf{87.0} &\cellcolor{gray!10}\textbf{84.9}&\cellcolor{gray!10}\textbf{95.5}&\cellcolor{gray!10}\textbf{50.6}&\cellcolor{gray!10} \textbf{75.4}\textsuperscript{\best{(+4.3)}
}\\ 
\cmidrule(lr){2-11} 

&TW AdaMerging++ \cite{yangadamerging}  &60.8& 56.9 &73.1& 83.4& 87.3& \textbf{82.4} &95.7& 50.1& 73.7\\ 
 &\cellcolor{gray!10}\textbf{w/ \method } (Ours) &\cellcolor{gray!10}\textbf{63.0 }&\cellcolor{gray!10}\textbf{63.2}&\cellcolor{gray!10}\textbf{78.5 }&\cellcolor{gray!10}\textbf{84.9}&\cellcolor{gray!10}\textbf{88.4} &\cellcolor{gray!10}79.1&\cellcolor{gray!10}\textbf{96.7}&\cellcolor{gray!10}\textbf{55.1}&\cellcolor{gray!10} \textbf{76.1}\textsuperscript{\best{(+2.4)}}\\ 
\cmidrule(lr){2-11} 
&LW AdaMerging \cite{yangadamerging} &64.5& 68.1& 79.2& \textbf{93.8}&\textbf{87.0}& \textbf{91.9}& \textbf{97.5}& \textbf{59.1}& \textbf{80.1} \\ 
&\cellcolor{gray!10} \textbf{w/ \method} (Ours)  &\cellcolor{gray!10}\textbf{68.0 }&\cellcolor{gray!10}\textbf{68.5}&\cellcolor{gray!10}\textbf{80.7}&\cellcolor{gray!10}88.3&\cellcolor{gray!10}80.0 &\cellcolor{gray!10}81.2&\cellcolor{gray!10}94.1&\cellcolor{gray!10}57.7& \cellcolor{gray!10}77.3\textsuperscript{\color[RGB]{255,0,0}{(-2.8)}}\\ 
\cmidrule(lr){2-11} 
&LW AdaMerging++ \cite{yangadamerging} &66.6& 68.3& \textbf{82.2}& \textbf{94.2} &\textbf{89.6} &\textbf{89.0}& \textbf{98.3}& \textbf{60.6}& \textbf{81.1}\\ 
 &\cellcolor{gray!10}\textbf{w/ \method} (Ours) &\cellcolor{gray!10}\textbf{68.9 }&\cellcolor{gray!10}\textbf{69.0}&\cellcolor{gray!10}81.0 &\cellcolor{gray!10}85.8&\cellcolor{gray!10} 82.2 &\cellcolor{gray!10}76.7&\cellcolor{gray!10}95.0&\cellcolor{gray!10}58.2&\cellcolor{gray!10}77.1\textsuperscript{\color[RGB]{255,0,0}{(-4.0)}}\\ 
\cmidrule(lr){2-11}

&PCB-MERGING \cite{du2024parameter} &63.8 &62.0 &77.1 &80.6& \textbf{87.5} &\textbf{78.5} &98.7 &58.4& 75.8  \\ 
&\cellcolor{gray!10} \textbf{w/ \method} (Ours)  &\cellcolor{gray!10}\textbf{68.6} &\cellcolor{gray!10}\textbf{67.7}&\cellcolor{gray!10}\textbf{80.7} &\cellcolor{gray!10}\textbf{83.0}&\cellcolor{gray!10}85.4&\cellcolor{gray!10}77.2&\cellcolor{gray!10}\textbf{96.2}&\cellcolor{gray!10}\textbf{60.0}&\cellcolor{gray!10} \textbf{77.3}\textsuperscript{\best{(+1.5)}}\\ 
\bottomrule

\\[-1.5ex]
\multirow{20}{*}{\raisebox{-15mm}{\rotatebox{90}{\textbf{ViT-L/14}}}}
&
Pretrained &66.8& 77.7& 71.0& 59.9& 58.4& 50.5& 76.3& 55.3& 64.5 \\ 

&Individual & 82.3 &92.4& 97.4& 100& 98.1 & 99.2& 99.7& 84.1& 94.2 \\ 


\cmidrule(lr){2-11}
&Weight Averaging \cite{wortsman2022model} & 72.1& 81.6& 82.6 &91.9& 78.2& 70.7& 97.1& 62.8& 79.6\\ 

&Fisher Merging \cite{matena2022merging} & 69.2& 88.6& 87.5& 93.5& 80.6& 74.8& 93.3 &70.0& 82.2 \\

&RegMean \cite{jindataless} & 73.3& 81.8& 86.1& 97.0& 88.0& 84.2& 98.5& 60.8& 83.7\\ 
&AdaMerging w/ Surgery \cite{yangrepresentation} &80.3 &90.8 &94.3& 98.2& 94.1 &98.7& 99.2& 82.5 &92.3 \\
\cmidrule(lr){2-11}

&Task Arithmetic \cite{ilharco2022editing} &73.9 &82.1 &86.6& 94.1& \textbf{87.9}& \textbf{86.7}&\textbf{ 98.9}& 65.6& 84.5 \\ 
&\cellcolor{gray!10}\textbf{w/ \method} (Ours) &\cellcolor{gray!10}\textbf{74.7} &\cellcolor{gray!10}\textbf{86.2 }&\cellcolor{gray!10}\textbf{89.3}&\cellcolor{gray!10}\textbf{94.6}&\cellcolor{gray!10}86.7&\cellcolor{gray!10}86.1&\cellcolor{gray!10} 98.8&\cellcolor{gray!10}\textbf{68.6}&\cellcolor{gray!10} \textbf{85.6}\textsuperscript{\best{(+1.1)}} \\ 
\cmidrule(lr){2-11}

&Ties \cite{yadav2023resolving} &\textbf{76.5}& 85.0& 89.3& \textbf{95.7}& \textbf{90.3} &83.3& 99.0& 68.8 &86.0\\
&\cellcolor{gray!10}\textbf{w/ \method} (Ours) &\cellcolor{gray!10}76.3&\cellcolor{gray!10}\textbf{87.4}&\cellcolor{gray!10}\textbf{90.9}&\cellcolor{gray!10}95.1&\cellcolor{gray!10}89.1 &\cellcolor{gray!10}\textbf{87.3}&\cellcolor{gray!10}\textbf{99.0}&\cellcolor{gray!10}\textbf{70.5}&\cellcolor{gray!10}\textbf{86.9}\textsuperscript{\best{(+0.9)}}\\
\cmidrule(lr){2-11}

&TW AdaMerging \cite{yangadamerging} &75.6 &83.5 &79.7 &90.3& 83.5 &96.5& 98.0& 67.3& 84.3 \\ 
&\cellcolor{gray!10}\textbf{w/ \method} (Ours) &\cellcolor{gray!10}\textbf{76.7} &\cellcolor{gray!10}\textbf{88.5 }&\cellcolor{gray!10}\textbf{87.7} &\cellcolor{gray!10}\textbf{96.3}&\cellcolor{gray!10}\textbf{90.7}&\cellcolor{gray!10}\textbf{97.6}&\cellcolor{gray!10}\textbf{98.7}&\cellcolor{gray!10}\textbf{75.6}& \cellcolor{gray!10}\textbf{89.0}\textsuperscript{\best{(+4.7)}} \\ 
\cmidrule(lr){2-11}
&TW AdaMerging++ \cite{yangadamerging} &76.6 &86.2 &85.6 &94.6& 89.7 &\textbf{96.8}& 98.2& 72.1& 87.5 \\ 

&\cellcolor{gray!10}\textbf{w/ \method} (Ours) &\cellcolor{gray!10}\textbf{78.0} &\cellcolor{gray!10}\textbf{89.3 }&\cellcolor{gray!10}\textbf{89.3} &\cellcolor{gray!10}\textbf{96.4}&\cellcolor{gray!10}\textbf{91.8 }&\cellcolor{gray!10}96.4&\cellcolor{gray!10}\textbf{98.8}&\cellcolor{gray!10}\textbf{77.2}& \cellcolor{gray!10}\textbf{89.7}\textsuperscript{\best{(+2.2)}} \\ 
\cmidrule(lr){2-11}

&LW AdaMerging \cite{yangadamerging} &79.0& 90.3 &90.8 &\textbf{96.2}& \textbf{93.4}& \textbf{98.0} &\textbf{99.0} &\textbf{79.9}& \textbf{90.8} \\ 
&\cellcolor{gray!10}\textbf{w/ \method} (Ours) &\cellcolor{gray!10}\textbf{79.0}&\cellcolor{gray!10}\textbf{90.3} &\cellcolor{gray!10}\textbf{91.6} &\cellcolor{gray!10}95.7& \cellcolor{gray!10}89.3&\cellcolor{gray!10} 96.4 &\cellcolor{gray!10}98.8 &\cellcolor{gray!10}77.6& \cellcolor{gray!10}89.8\textsuperscript{\color[RGB]{255,0,0}{(-1.0)}} \\
\cmidrule(lr){2-11}

&LW AdaMerging++ \cite{yangadamerging} &\textbf{79.4}&\textbf{90.3}& 91.6& \textbf{97.4}& \textbf{93.4}& \textbf{97.5}&\textbf{99.0}& \textbf{79.2}& \textbf{91.0} \\
&\cellcolor{gray!10} \textbf{w/ \method} (Ours) &\cellcolor{gray!10}79.2&\cellcolor{gray!10}90.1&\cellcolor{gray!10}\textbf{92.4}&\cellcolor{gray!10}96.4&\cellcolor{gray!10}89.1&\cellcolor{gray!10}94.9&\cellcolor{gray!10}98.8&\cellcolor{gray!10}76.6&\cellcolor{gray!10}89.7\textsuperscript{\color[RGB]{255,0,0}{(-1.3)}} \\ 
\cmidrule(lr){2-11}

&PCB-MERGING \cite{du2024parameter}  & 76.2& 86.0 &89.6& \textbf{95.9}& \textbf{89.9} &\textbf{92.3}& \textbf{99.2}& 71.4 & 87.6\\ 
&\cellcolor{gray!10}\textbf{w/ \method} (Ours) & \cellcolor{gray!10}\textbf{76.8}&\cellcolor{gray!10}\textbf{88.2} &\cellcolor{gray!10}\textbf{91.1}&\cellcolor{gray!10}95.8&\cellcolor{gray!10}89.2 &\cellcolor{gray!10}92.1&\cellcolor{gray!10}99.1&\cellcolor{gray!10}\textbf{72.7} &\cellcolor{gray!10}\textbf{88.1}\textsuperscript{\best{(+0.5)}}\\ 
\bottomrule
\end{tabular}
}
\end{table*}

\begin{table*}[htbp]
\centering
\caption{Performance comparison for ID  datasets on ViT-H/14.}
\label{ViT-H-14-ID}
\setlength{\tabcolsep}{1mm}
\resizebox{\textwidth}{!}{%
\begin{tabular}{cccccccccc}
\toprule
 \multirow{1}{*}{\textbf{Method (↓)}} & \textbf{SUN397} & \textbf{Cars} & \textbf{RESISC45} & \textbf{EuroSAT} & \textbf{SVHN} & \textbf{GTSRB}& \textbf{MNIST}& \textbf{DTD}& \textbf{Avg.} \\ 
\midrule

Pretrained  &73.8 &93.4 & 75.7& 73.1&52.5 &58.4 &72.8 &67.8 & 70.9 \\ 

 Individual &81.7 &95.1 &97.5 & 99.3& 98.1& 99.3& 99.7& 83.8&94.3  \\

\midrule
Weight Averaging \cite{wortsman2022model} & 76.2&94.5 &85.8 &92.7 &84.6 & 79.8&97.5 & 70.1&85.2 \\ 
\midrule

Task Arithmetic \cite{ilharco2022editing} &77.0 &94.5 & 89.6& 94.7&91.3 &\textbf{91.0} & \textbf{99.3}&69.6 &88.4 \\ 
\rowcolor{gray!10} \textbf{w/ \method} (Ours) &\textbf{78.0} &\textbf{95.2} & \textbf{92.0}&\textbf{96.4} & \textbf{92.1}& 90.4&99.1 & \textbf{71.8}&\textbf{89.4}\textsuperscript{\best{(+1.0)}}\\ 
\midrule
Ties \cite{yadav2023resolving} &77.7 &94.8 &90.1&94.3 &91.2 &86.8 &99.0 &70.1 &88.0 \\
 \rowcolor{gray!10}\textbf{w/ \method} (Ours) &\textbf{78.9} &\textbf{95.2}&\textbf{92.6}&\textbf{96.3} &\textbf{93.8} &\textbf{91.2} &\textbf{99.3} &\textbf{71.3} &\textbf{89.8}\textsuperscript{\best{(+1.8)}} \\
\midrule
PCB-MERGING \cite{du2024parameter} &77.8 &94.6 &90.6 &96.2 &93.5& 95.1 &\textbf{99.4} & 76.4 &90.5 \\ 
\rowcolor{gray!10}\textbf{w/ \method} (Ours)&\textbf{78.9} &\textbf{95.3} &\textbf{92.8} & \textbf{96.7}&\textbf{94.0} &\textbf{95.2} &99.3&\textbf{78.7} & \textbf{91.4}\textsuperscript{\best{(+1.8)}} \\ 
\bottomrule
\end{tabular}
}
\end{table*}

\subsection{Performance comparison on ViT-H/14}\label{sub:ViT-H-14}

Tab.\ref{ViT-H-14} demonstrates the comparative evaluation of our method against multiple baselines across ID and OOD datasets, implemented on the ViT-H/14 architecture. It can be seen that although the ViT-H/14 model is already very large, there is still a gap in the generalization performance between the pretrained model and the individual fine-tuned model on the OOD datasets. Therefore, our method still works. Experimental results reveal that our method yields significant performance gains over existing techniques, with H-score improvements of 1.4\%, 0.9\%, and 1.3\% over Task Arithmetic, Ties, and PCB, respectively, when combined with these approaches.

\begin{table*}[htbp]
\centering
\caption{Performance comparison for ID and OOD datasets on ViT-H/14.}
\label{ViT-H-14}
\tiny
\setlength{\tabcolsep}{0.5mm}
\resizebox{\textwidth}{!}{%
\begin{tabular}{@{}ccccccccccccccccc@{}}
\toprule
  \multirow{1}{*}{\textbf{Dataset (→)}}  &  \multicolumn{1}{c}{\textbf{In Domain}} & \multicolumn{14}{c}{\textbf{Out of Domain}} &\multirow{2}{*}{ \textbf{H-score}} \\ 
  \cmidrule(lr){2-2} \cmidrule(lr){3-16}  
 \multirow{1}{*}{\textbf{Method (↓)}} & \textbf{Avg.} & \textbf{C10} & \textbf{C100} & \textbf{EMN} & \textbf{FMN} & \textbf{FER}& \textbf{F102}& \textbf{F101}& \textbf{KMN}& \textbf{OxP} & \textbf{PCA} & \textbf{RSS} & \textbf{STL}& \textbf{Ima} & \textbf{Avg.}  \\ 
\midrule
Pretrained & 70.9  & 97.4  & 84.7 & 14.1 &79.0 & 36.2 & 80.1 & 92.2 & 11.7 & 94.5  & 51.0  & 64.0& 98.5 &77.9 & 67.8&  69.3  \\ 

Individual & 94.3& 91.7 &72.6  & 18.7 &  74.2& 37.4 & 77.7&88.8 &10.1  & 93.4& 52.1 & 63.9 &97.4  &75.0   & 65.6  &   77.4 \\

\midrule
Weight Averaging \cite{wortsman2022model}
&85.2 & 97.1 & 83.0& 21.5 &79.7 &38.2 &79.7 & 91.5 &9.9 & 94.3& 51.3& 63.9 &98.6 & 77.7& 68.2 &  75.8 \\ 

\midrule

Task Arithmetic \cite{ilharco2022editing} 
& 88.4& 94.4  &  74.8 &  \textbf{28.9} & 78.7 &38.1 & 77.9 &  88.0 & 9.4 & 93.3 &\textbf{52.3} &62.9 & 97.6 &75.0 &67.0& 76.2   \\
\rowcolor{gray!10}\textbf{ w/ \method} (Ours)& \textbf{89.4} &\textbf{96.4}&\textbf{81.4}& 28.6&\textbf{79.9 }&\textbf{39.0}& \textbf{79.3}& \textbf{91.6}& \textbf{9.7 }&\textbf{94.3} & 51.7 &\textbf{64.1}  & \textbf{98.5} &   \textbf{77.7} &\textbf{68.6} \textsuperscript{\best{(+1.6)}} & \textbf{77.6} \textsuperscript{\best{(+1.4)} } \\
\midrule


Ties \cite{yadav2023resolving} & 88.9 & 95.0  &76.4 & 27.0 & 78.8&38.4&78.6 &89.2  & 9.2 & 93.8 & \textbf{51.9} &64.1 &98.0 &76.0 &67.4 & 76.7 \\ 
\rowcolor{gray!10}\textbf{ w/ \method} (Ours)
& \textbf{89.8}& \textbf{96.0} &\textbf{ 79.9} &\textbf{ 27.7} & \textbf{79.5}
 & \textbf{38.8} &\textbf{ 79.1} & \textbf{91.4} & \textbf{9.5 }& \textbf{94.3} & 51.7 &\textbf{64.1} & \textbf{98.5} & \textbf{77.5} & \textbf{68.3} \textsuperscript{\best{(+0.9)}}& \textbf{77.6}  \textsuperscript{\best{(+0.9)}}  \\
\midrule

PCB-MERGING \cite{du2024parameter} &90.5&  93.5 &  72.7
  & 27.8 & 78.8 &37.3 & 76.8 & 87.9  & 9.4  & 93.3 & \textbf{52.4} & 63.3 & 97.7 & 75.0 &66.6 &  76.7 \\
\rowcolor{gray!10}\textbf{ w/ \method} (Ours)& \textbf{91.4} & \textbf{95.3} & \textbf{78.0} & \textbf{28.2 }&\textbf{79.3} & \textbf{38.6 }&\textbf{ 78.6 } &  \textbf{91.1}&  \textbf{ 9.4 } &\textbf{94.3}  &51.9 & \textbf{64.1} & \textbf{98.4} &\textbf{77.3}  & \textbf{68.0} \textsuperscript{\best{(+1.4)}} & \textbf{78.0} \textsuperscript{\best{(+1.3)}} \\
\bottomrule
\end{tabular}
}
\end{table*}

\subsection{More ablation study details}
Tab.\ref{tab:ablation_ID} and Tab.\ref{tab:ablation_OOD} provides a detailed summary of the experimental results from our comprehensive ablation studies performed on the ID dataset and OOD dataset.

\begin{table}[H]
  \centering
  \scriptsize
  \setlength{\extrarowheight}{1.0pt}
  \setlength{\tabcolsep}{1.1mm}
  \rowcolors{2}{gray!10}{white}
  \caption{The ablation study of \method on the ViT-B/32 regarding the ID dataset.}
  \label{tab:ablation_ID}
  \begin{adjustbox}{width=\textwidth}
  \begin{tabular}{l ccc ccccccccc}
    \toprule
    \rowcolor{gray!30}
    Method & \multicolumn{3}{c}{Components} & \multicolumn{9}{c}{Performance (\%)} \\
    \cmidrule(lr){2-4}\cmidrule(lr){5-13}
           & Salience Score & $\lor$ & Scale & SUN397& Cars& RESISC45& EuroSAT& SVHN& GTSRB& MNIST& DTD& Avg. \\
    \midrule
   
           & \xmark & \xmark & -      & 55.2& 54.9& 66.7& 78.9& 80.2& 69.7& 97.3& 50.4& 69.1     \\
           & \cmark & \xmark & -      & 68.1& 65.7& 76.0& 77.4& 69.2& 61.2& 87.1& 54.4& 69.9   \\
           & \xmark & \cmark & -      & 64.3& 62.1& 65.0& 57.9& 35.9& 36.6& 47.6& 45.9& 51.9  \\
      \multirow{-4}{*}{Task Arithmetic \cite{ilharco2022editing}}      & \cmark & \cmark & -      & 66.8& 66.0& 77.2& 79.1& 78.3& 68.6& 92.3& 54.4& 72.8\\
    \midrule
   
           & \xmark & \xmark & \xmark & 64.5 &68.1& 79.2& 93.8& 87.0& 91.9& 97.5 &59.1& 80.1     \\
           & \cmark & \cmark & \xmark & 66.0& 67.7& 82.0& 91.4& 87.5& 87.7& 97.3& 59.0& 79.8     \\
           & \cmark & \xmark & \cmark & 68.6& 66.6& 79.8& 83.7& 79.1& 75.4& 91.4& 57.3& 75.2     \\
           & \xmark & \cmark & \cmark & 64.5& 62.5& 67.2& 62.4& 38.3& 39.7& 47.8& 46.7& 53.6     \\
         \multirow{-5}{*}{LW AdaMerging \cite{yangadamerging}}   & \cmark & \cmark & \cmark & 68.0& 68.5& 80.7& 88.3& 80.0& 81.2& 94.1& 57.7& 77.3 \\
    \bottomrule
  \end{tabular}
  \end{adjustbox}
\end{table}

\begin{table}[H]
  \centering
  \scriptsize
  \setlength{\extrarowheight}{1.0pt}
  \setlength{\tabcolsep}{0.8mm}
  \rowcolors{2}{gray!10}{white}
  \caption{The ablation study of \method on the ViT-B/32 regarding the OOD dataset.}
  \label{tab:ablation_OOD}
  \begin{adjustbox}{width=\textwidth}
  \begin{tabular}{l ccc cccccccccccccc}
    \toprule
    \rowcolor{gray!30}
    Method & \multicolumn{3}{c}{Components} & \multicolumn{14}{c}{Performance (\%)} \\
    \cmidrule(lr){2-4}\cmidrule(lr){5-18}
           & Salience Score & $\lor$ & Scale & C10& C100 &EMN &FMN& FER& F102& F101& KMN& OxP& PCA& RSS& STL& Ima& Avg. \\
    \midrule
   
           & \xmark & \xmark & -      &76.3& 41.9& 28.5& 63.9& 26.3& 49.4& 55.8 &9.0& 75.4& 54.0& 53.4& 87.8& 45.0&51.3    \\
           & \cmark & \xmark & -      & 89.4 &62.6& 28.5& 66.1 &38.6& 64.8 &81.2& 7.2 & 86.5& 59.6& 59.2& 96.9& 62.2& 61.8    \\
           & \xmark & \cmark & -      & 90.3& 64.4& 17.9& 63.8& 39.4& 66.5& 82.7& 9.4& 86.9& 59.0& 57.0& 97.1& 62.7& 61.3    \\
      \multirow{-4}{*}{Task Arithmetic \cite{ilharco2022editing}}      & \cmark & \cmark & -& 88.3& 60.1& 30.4& 65.9& 36.5& 62.9& 79.9& 7.4& 85.4& 59.5& 59.0& 96.4& 60.6& 60.9 \\
    \midrule
   
           & \xmark & \xmark & \xmark & 82.1& 50.7& 28.7& 63.2& 37.8& 58.3& 73.8& 8.8& 82.4& 57.3& 58.0& 94.3& 54.6& 57.7     \\
           & \cmark & \cmark & \xmark & 84.1& 52.0& 30.1& 64.1& 37.1& 59.6& 76.5& 8.3& 83.3& 57.6& 58.1& 95.4& 56.4& 58.7     \\
           & \cmark & \xmark & \cmark & 88.4& 60.7& 30.6& 65.8& 38.7& 63.5& 80.4& 7.2& 86.0& 59.4& 58.8& 96.7& 61.3& 61.3     \\
           & \xmark & \cmark & \cmark & 90.4& 64.5& 19.5& 64.5& 39.0& 65.1& 82.0& 9.2& 85.5& 57.4& 56.0& 97.0& 61.3& 60.9      \\
         \multirow{-5}{*}{LW AdaMerging \cite{yangadamerging}}   & \cmark & \cmark & \cmark & 88.8& 61.3& 30.6& 66.1& 38.4& 64.3& 80.6&  7.6& 86.3 &60.1&  59.4& 96.6& 61.2& 61.7  \\
    \bottomrule
  \end{tabular}
  \end{adjustbox}
\end{table}

\subsection{Comparison with OOD methods for fine-tuned Models}
\label{sec:OOD exc}
Tab.\ref{table:OOD and Pruning ID} and Tab.\ref{table:OOD_and_Pruning_OOD} provide a detailed summary of the experimental results comparing OOD methods for fine-tuned models on the ID dataset and OOD dataset. Following Wortsman et al. \cite{wortsman2022robust}, we set the mixing coefficient $\alpha=0.5$ for WiSE-FT. For the Model Stock, we employed the fine-tuning code from Ilharco et al. \cite{ilharco2022editing} and set the random seed to 10 to obtain the second set of fine-tuned models.
We set the search range for $\lambda$ in the Task Arithmetic to $(0,1]$
 with a step size of 0.1 base on Ilharco et al. \cite{ilharco2022editing, wang2024lines}.

\begin{table}[H]
\centering
\caption{ \small{Comparison with other OOD methods (above) and purning methods (below) on  ViT-B/32 on the ID dataset.}}
\label{table:OOD and Pruning ID}
\scriptsize
\setlength{\tabcolsep}{2pt}
\resizebox{\textwidth}{!}{
\begin{tabular}{@{}c|cccccccccc@{}}
\toprule
 \multicolumn{2}{c}{\textbf{Dataset (→)}}  &  \multirow{2}{*}{\textbf{SUN397}} & \multirow{2}{*}{\textbf{Cars}}&\multirow{2}{*}{\textbf{RESISC4}}&\multirow{2}{*}{\textbf{EuroSAT}}&\multirow{2}{*}{\textbf{SVHN}}&\multirow{2}{*}{\textbf{GTSRB}}&\multirow{2}{*}{\textbf{MNIST}}&\multirow{2}{*}{\textbf{DTD}} &\multirow{2}{*}{\textbf{Avg.}} \\
\multicolumn{2}{c}{\textbf{Method (↓)}} \\ 
\midrule
Baseline&Task Arithmetic \cite{ilharco2022editing} 
& 55.2 & 54.9& 66.7& 78.9& 80.2& 69.7& 97.3& 50.4& 69.1   \\  
\midrule
\rowcolor{lightgray!10}  & w/ WiSE-FT \cite{wortsman2022robust}
& 60.6 & 59.0 &70.2 & 79.4 &78.2 & 68.6 & 96.1 &51.5 & 70.5   \\ 

 \rowcolor{lightgray!10} &  Model Stock \cite{jang2024model}
& 60.7 &65.0 &75.0 & 81.7 & 80.0 & 68.5 & 95.2 & 49.6 & 72.0   \\ 
\rowcolor{lightgray!10} \multirow{-3}{*}{OOD}   &w/ LiNeS  \cite{wang2024lines}
& 63.9& 63.9 & 75.1 & 85.6 & 79.4 & 72.2 & 96.2 &56.5 & 74.1   \\

\midrule


 \rowcolor{lightgray!30}& w/ DARE  \cite{yu2024language}
& 64.6 & 58.7 &  60.5 & 47.4 & 32.1 &33.6 &51.0 &44.8 & 49.1   \\ 
\rowcolor{lightgray!30}& w/ MWP \cite{sanh2020movement}
& 62.6 & 61.6 &72.1 &78.1 & 80.0 &69.9 &96.6 &53.0 &71.7   \\ 
\rowcolor{lightgray!30} & w/ random mask
& 66.2 & 59.9 & 61.3 & 48.6 & 32.1 & 32.3 & 48.2 & 44.5 &49.1  \\ 
\rowcolor{lightgray!30} \multirow{-4}{*}{Pruning} & w/ absolute  value 
& 64.6 &61.0 & 60.7 & 47.1& 32.0 & 32.8 & 48.4 & 44.4 & 48.9 \\ 
\midrule
\textbf{Ours}&\textbf{ w/ LwPT} & 66.8& 66.0& 77.2& 79.1& 78.3& 68.6& 92.3& 54.4& 72.8 \\ 
\bottomrule
\end{tabular}
}
\end{table}

\begin{table}[H]
\centering
\caption{ \small{Comparison with other OOD methods (above) and purning methods (below) on  ViT-B/32 on the OOD dataset.}}
\label{table:OOD_and_Pruning_OOD}
\scriptsize
\setlength{\tabcolsep}{2pt}
\resizebox{\textwidth}{!}{
\begin{tabular}{@{}c|ccccccccccccccc@{}}
\toprule
 \multicolumn{2}{c}{\textbf{Dataset (→)}}  &  \multirow{2}{*}{\textbf{C10}} & \multirow{2}{*}{\textbf{C100}}&\multirow{2}{*}{\textbf{EMN}}&\multirow{2}{*}{\textbf{FMN}}&\multirow{2}{*}{\textbf{FER}}&\multirow{2}{*}{\textbf{F102}}&\multirow{2}{*}{\textbf{F101}}&\multirow{2}{*}{\textbf{KMN}} &\multirow{2}{*}{\textbf{OxP}} &\multirow{2}{*}{\textbf{PCA}} &\multirow{2}{*}{\textbf{RSS}} &\multirow{2}{*}{\textbf{STL}} &\multirow{2}{*}{\textbf{Ima}} &\multirow{2}{*}{\textbf{Avg.}}   \\
\multicolumn{2}{c}{\textbf{Method (↓)}} \\ 
\midrule
Baseline&Task Arithmetic \cite{ilharco2022editing} 
&76.3 &41.9& 28.5& 63.9& 26.3& 49.4& 55.8& 9.0& 75.4& 54.0 &53.4& 87.8& 45.0& 51.3   \\  
\midrule
\rowcolor{lightgray!10}  & w/ WiSE-FT \cite{wortsman2022robust}
&81.6 &49.9 &29.6 &64.5 &30.3 & 55.2 & 65.9 & 8.4 & 81.1 &56.7 & 54.2 & 91.4 & 52.0 &55.4   \\ 

 \rowcolor{lightgray!10} &  Model Stock \cite{jang2024model}
& 86.6 &57.7 & 30.9 &65.1& 32.1 &60.8 & 74.9 & 8.7 & 84.2 & 57.0 & 54.4 &  94.9 & 56.5 & 58.7 \\ 
\rowcolor{lightgray!10} \multirow{-3}{*}{OOD}   &w/ LiNeS  \cite{wang2024lines}
& 84.4 &54.9 &30.3 &65.4 &33.6 &60.4 &  73.7 & 8.4 & 83.6 & 58.8 & 51.4 & 93.9 & 55.6 &58.0\\

\midrule


 \rowcolor{lightgray!10}& w/ DARE  \cite{yu2024language}
& 88.7 & 64.7& 20.8 &64.1 &38.0 &65.9 & 82.3&8.8 & 87.0 & 61.7 & 58.2 &96.9 & 63.1 &61.5   \\ 
\rowcolor{lightgray!10}& w/ MWP \cite{sanh2020movement}
& 83.7 & 52.1 & 27.8& 64.4& 32.0 & 56.1 & 69.2 & 8.2 & 81.3 & 57.5  &54.2   &92.9  & 53.6  & 56.4  \\ 
\rowcolor{lightgray!10} & w/ random mask
& 89.6 & 64.9 & 18.1 & 63.9 & 38.8& 66.5 & 82.9 & 9.4& 87.7 & 60.2 & 59.2 & 97.0 & 63.6 & 61.7 \\ 
\rowcolor{lightgray!10} \multirow{-4}{*}{Pruning} & w/ absolute  value 
& 89.9& 64.6 & 17.5 & 63.7 & 38.8 & 66.6 & 82.9 & 9.4 & 87.4 & 60.2& 57.9 & 97.2 & 63.3 & 61.5\\ 
\midrule
\textbf{Ours}&\textbf{ w/ LwPT} & 88.3& 60.1& 30.4& 65.9& 36.5& 62.9& 79.9& 7.4& 85.4& 59.5& 59.0& 96.4& 60.6& 60.9 \\ 
\bottomrule
\end{tabular}
}
\end{table}

\subsection{Comparison with other pruning methods}
\label{sec:Pruning}

Tab.\ref{table:OOD and Pruning ID} and Tab.\ref{table:OOD_and_Pruning_OOD} provide a detailed summary of the experimental results comparing other pruning methods on the ID dataset and OOD dataset. In which,  for the DARE and random mask methods, we select the best results obtained from three runs with different random seeds.

\subsection{The analysis of layer-wise pruning and parameter-wise pruning}
\label{sec:Layer_Parameter}
\begin{wraptable}[8]{r}{6.5cm}
\vspace{-1.2em}
\centering
\caption{ \small{The comparison of layer-wise pruning and parameter-wise pruning on ViT-B/32.}}
\label{table:Layer and parameter}
\scriptsize
\setlength{\tabcolsep}{2pt}
\begin{tabular}{@{}c|ccc@{}}
\toprule
 
\multicolumn{2}{c}{\textbf{Method (↓)}}&\multirow{1}{*}{\textbf{H-score}} & \multirow{1}{*}{\textbf{Time consumption}}  \\ 
\midrule
Baseline&Task Arithmetic \cite{ilharco2022editing} 
& 58.9&  8.3   \\  
\midrule

\rowcolor{lightgray!10} &\textbf{ w/ PwPTV} & 61.7 &135.4s    \\
 \rowcolor{lightgray!10} \multirow{-2}{*}{\textbf{Ours}} &\textbf{ w/ \method} & 66.3 &9.0s    \\ 
\bottomrule
\end{tabular}
\end{wraptable}
To investigate the distinctions between layer-wise pruning (LwPTV), which computes significance scores layer by layer, and parameter-wise pruning (PwPTV), which computes significance scores for individual parameters, under identical experimental setups, we conducted a comparative analysis, with a focus on performance and time consumption, based on the Task Arithmetic method on the ViT-B/32 architecture. The results are delineated in Tab.\ref{table:Layer and parameter}. The detailed results of the two methods on the ID and OOD datasets are presented in Tab.\ref{table:ID_Layer_and_Par} and Tab.\ref{table:OOD_Layer_and_Par}, respectively.
From this, it can be observed that LwPTV achieves better performance with less time consumption. This performance improvement can be attributed to LwPTV, which aligns the representations of OOD data generated by the merged model more closely with those produced by the pretrained model, thereby significantly enhancing OOD performance.

\begin{table}[H]
\centering
\caption{ \small{The comparison of layer-wise pruning and parameter-wise pruning on ID datasets.}}
\label{table:ID_Layer_and_Par}
\scriptsize
\setlength{\tabcolsep}{2pt}
\resizebox{\textwidth}{!}{
\begin{tabular}{@{}c|cccccccccc@{}}
\toprule
 \multicolumn{2}{c}{\textbf{Dataset (→)}}  &  \multirow{2}{*}{\textbf{SUN397}} & \multirow{2}{*}{\textbf{Cars}}&\multirow{2}{*}{\textbf{RESISC4}}&\multirow{2}{*}{\textbf{EuroSAT}}&\multirow{2}{*}{\textbf{SVHN}}&\multirow{2}{*}{\textbf{GTSRB}}&\multirow{2}{*}{\textbf{MNIST}}&\multirow{2}{*}{\textbf{DTD}} &\multirow{2}{*}{\textbf{Avg.}} \\
\multicolumn{2}{c}{\textbf{Method (↓)}} \\ 
\midrule
Baseline&Task Arithmetic \cite{ilharco2022editing} 
& 55.2 & 54.9& 66.7& 78.9& 80.2& 69.7& 97.3& 50.4& 69.1   \\  
\midrule
\rowcolor{lightgray!10}  & w/  PwPTV
& 59.8 & 59.0 &70.6 & 79.0 &81.9 & 72.2 & 97.2 &52.3& 71.5   \\ 

 \rowcolor{lightgray!10} \multirow{-2}{*}{Ours}  &  w/ \method  & 66.8 &66.0 &77.2 &79.1& 78.3 &68.6 &92.3& 54.4& 72.8\\
\bottomrule
\end{tabular}
}
\end{table}

\begin{table}[H]
\centering
\caption{ \small{The comparison of layer-wise pruning and parameter-wise pruning on OOD datasets.}}
\label{table:OOD_Layer_and_Par}
\scriptsize
\setlength{\tabcolsep}{2pt}
\resizebox{\textwidth}{!}{
\begin{tabular}{@{}c|ccccccccccccccc@{}}
\toprule
 \multicolumn{2}{c}{\textbf{Dataset (→)}}  &  \multirow{2}{*}{\textbf{C10}} & \multirow{2}{*}{\textbf{C100}}&\multirow{2}{*}{\textbf{EMN}}&\multirow{2}{*}{\textbf{FMN}}&\multirow{2}{*}{\textbf{FER}}&\multirow{2}{*}{\textbf{F102}}&\multirow{2}{*}{\textbf{F101}}&\multirow{2}{*}{\textbf{KMN}} &\multirow{2}{*}{\textbf{OxP}} &\multirow{2}{*}{\textbf{PCA}} &\multirow{2}{*}{\textbf{RSS}} &\multirow{2}{*}{\textbf{STL}} &\multirow{2}{*}{\textbf{Ima}} &\multirow{2}{*}{\textbf{Avg.}}   \\
\multicolumn{2}{c}{\textbf{Method (↓)}} \\ 
\midrule
Baseline&Task Arithmetic \cite{ilharco2022editing} 
&76.3 &41.9& 28.5& 63.9& 26.3& 49.4& 55.8& 9.0& 75.4& 54.0 &53.4& 87.8& 45.0& 51.3   \\  
\midrule

 \rowcolor{lightgray!10} &  w/ PwPTV 
&  80.4 &46.6 & 27.8 &64.5& 29.0 &53.0 & 63.9 & 8.4 & 79.0 & 56.8 & 54.1 &  90.8 & 50.1 & 54.2 \\ 
\rowcolor{lightgray!10} \multirow{-2}{*}{Ours}   &w/  \method
& 88.3& 60.1& 30.4& 65.9& 36.5& 62.9& 79.9& 7.4& 85.4& 59.5& 59.0& 96.4& 60.6& 60.9\\ 
\bottomrule
\end{tabular}
}
\end{table}

\begin{wrapfigure}[17]{r}{7.5cm}
\vspace{-2.6em}
    \centering
\includegraphics[width=0.60\textwidth]{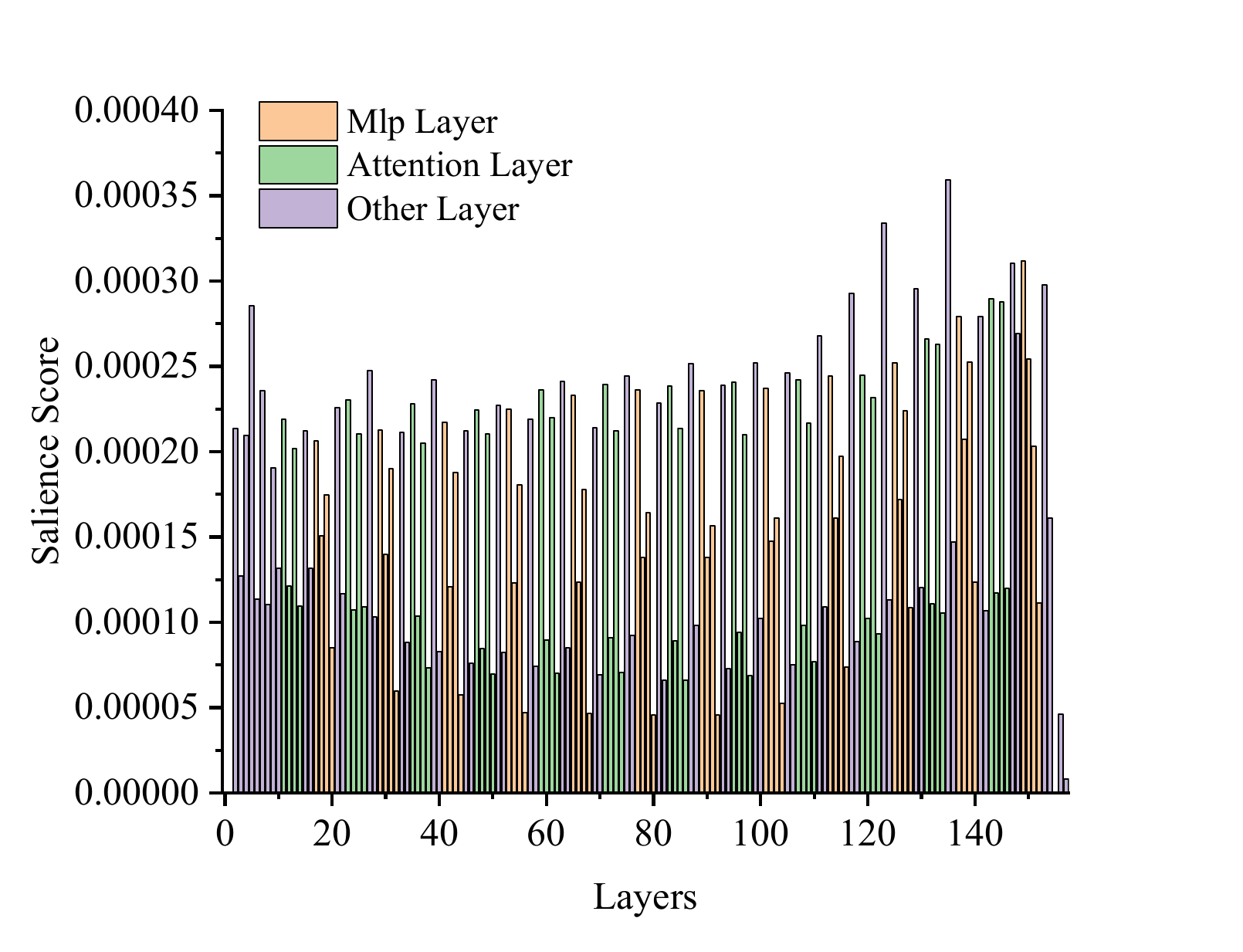}
\vspace{-2.0em}
  \caption{\small{ The salience score of the task vectors with different component on ViT-B/32.}}
\label{fig:ViT-salience_Component}
\end{wrapfigure}

\subsection{Neural network component analysis}
\label{sec:componet}
To investigate whether our method primarily prunes task vectors specific to certain structures, such as parameters in MLP, whose pre-trained parameters inherently have stronger generalization capabilities, leading to the effectiveness of LwPTV. We conducted a more in-depth analysis of ours by categorizing the layers of ViT into three types: \textbf{Attention Layer}, \textbf{MLP Layer}, and \textbf{Other Layer} (including position embedding, layer normalization, etc.). We then analyzed their  salience scores of the task vectors for SUN397. As shown in Fig. \ref{fig:ViT-salience_Component} the salience scores  do not reflect this trend, suggesting that ours does not selectively filter out specific types of layers. Therefore, the effectiveness of ours lies in pruning the redundant parameters, instead of pruning certain layer types.

\subsection{\method aligns the representations between the merged model and the pretrained model}

\label{sec:representations}

\begin{figure}[H]
  \centering
  \begin{subfigure}[b]{0.48\textwidth}
    \includegraphics[width=\linewidth]{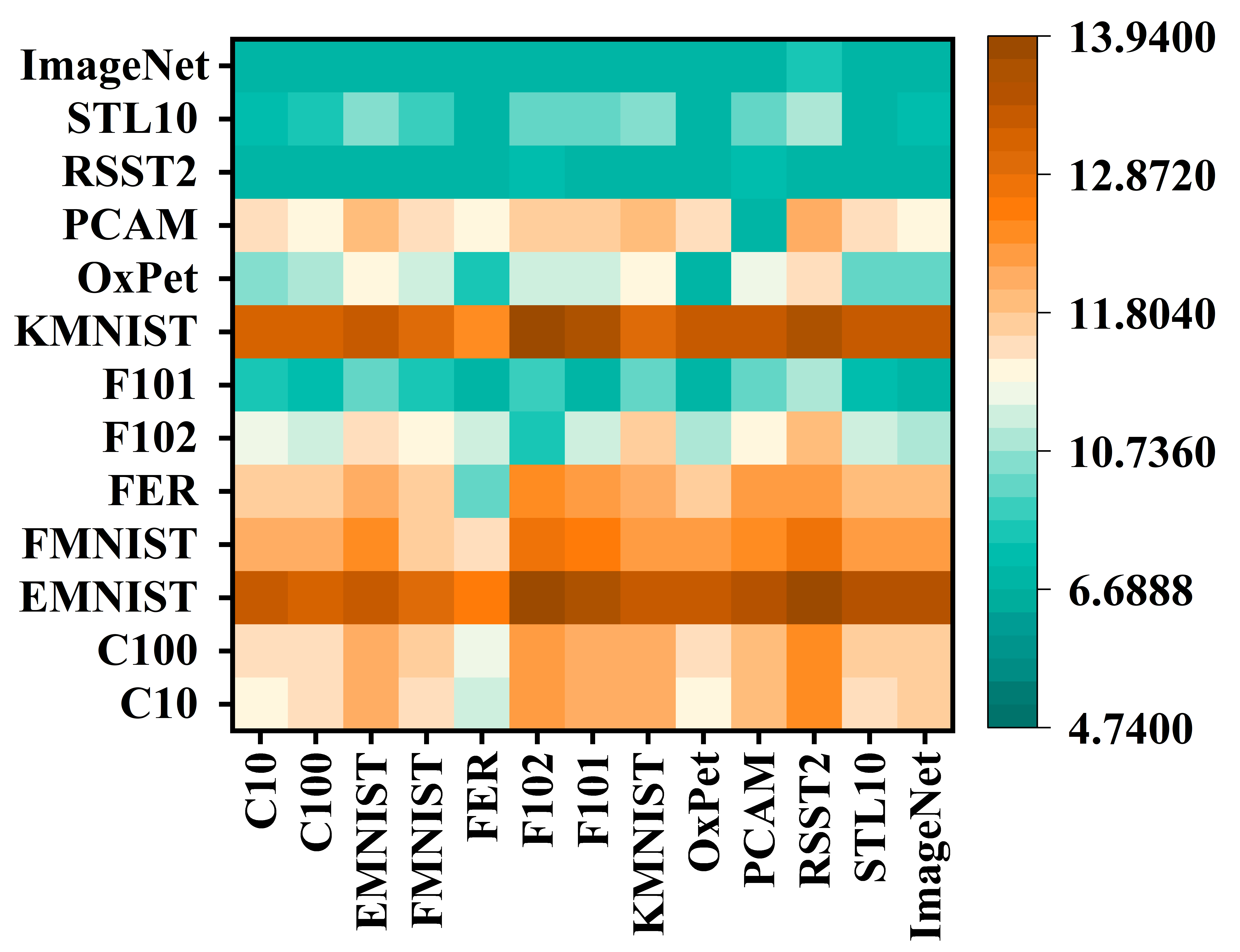}
    \caption{Task Arithmetic \textit{vs.} Pretrained model}
    \label{fig:no_pruning_distance}
  \end{subfigure}
  \hfill
  \begin{subfigure}[b]{0.48\textwidth}
    \includegraphics[width=\linewidth]{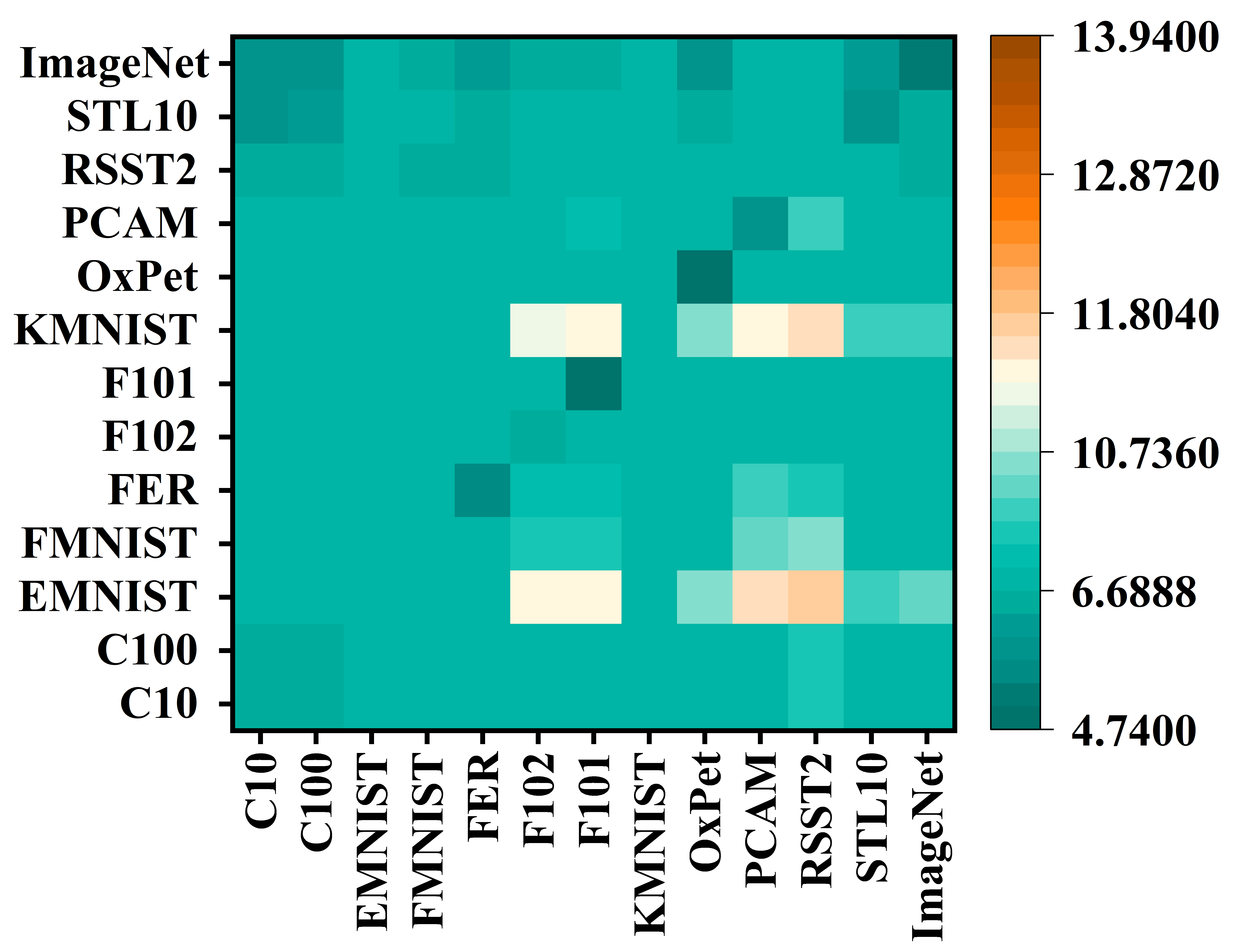}
    \caption{Task Arithmetic w/ \method \textit{vs.} Pretrained model}
    \label{fig:pruning_distance}
  \end{subfigure}

  \vskip\baselineskip  

   \caption{Visualization of $\ell_2$ distance between merged  model representations and pre-trained model representations.}

  \label{fig:rep_l2_distance}
\end{figure}

 To investigate how our method \method influences the merged model in generating representations for OOD data, we compute two sets of $\ell_2$ distances: (1) Task Arithmetic \textit{vs.} Pretrained model (2) Task Arithmetic w/ \method \textit{vs.} Pretrained model. The results are shown in Fig. \ref{fig:rep_l2_distance}. It can be seen that \method aligns the representations between the merged model and the pretrained model. This observation is consistent with our goal and further supports the notion that our method improves OOD performance.

\subsection{Storage overhead}
\label{sec:Storage Overhead}
\method can effectively diminish the storage requirements for task vectors for each merging method. As shown in Tab.\ref{storage_cost}, we report the storage cost  derived from various model merging techniques applied to the ViT-B/32 architecture. It can be seen that, \method can significantly reduce the storage cost of the merged model in various methods. Since both Ties and PCB prune the parameters, applying ours to them further lead to smaller storage cost.

\begin{table}[!htbp]
\centering
\scriptsize
\caption{The storage cost of various model merging methods on ViT-B/32 and ViT-L/14.}
\label{storage_cost}

\setlength{\tabcolsep}{0.5mm}
\begin{adjustbox}{width=\textwidth}
\begin{tabular}{@{}cccccccccc@{}}
\toprule
  \multirow{2}{*}{\textbf{Model Architecture}}& \multirow{1}{*}{\textbf{Method (→)}}& \multirow{2}{*}{ \textbf{Pretrained model}}  &  \multirow{2}{*}{ \textbf{Task Arithmetic }} &\multirow{2}{*}{ \textbf{Ties }}&\multirow{2}{*}{ \textbf{TW AdaMerging }}& \multirow{2}{*}{ \textbf{TW AdaMerging++ }} & \multirow{2}{*}{ \textbf{LW AdaMerging}}& \multirow{2}{*}{ \textbf{LW AdaMerging++ }}  &\multirow{2}{*}{ \textbf{PCB}} \\  
 &\multirow{1}{*}{\textbf{w/\method (↓)}}   \\ 
\midrule

\multirow{2}{*}{\textbf{ViT-B/32}}& \xmark &432.77 MB&432.77 MB & 285.25 MB&432.77 MB &285.25 MB&432.77 MB & 285.25 MB& 123.39 MB\\

 &\cmark &-&226.75 MB & 220.03 MB&226.75 MB & 220.03 MB&226.75 MB &220.03 MB & 115.23 MB\\
 \midrule

 \multirow{2}{*}{\textbf{ViT-L/14}}& \xmark & 1306.77 MB &1306.77 MB & 965.18 MB  &1306.77 MB & 965.18 MB
 & 1306.77 MB&965.18 MB &397.26 MB  \\

 &\cmark &-&836.35 MB &788.08 MB &836.35 MB &788.08 MB &836.35 MB &788.08 MB & 377.21 MB \\
\bottomrule
\end{tabular}
\end{adjustbox}
\end{table}

\subsection{Time complexity analysis }
\label{sec:Time complexity analysis}

\begin{wraptable}[7]{r}{4cm}
\vspace{-1.3em}
\centering
\caption{ \small{Time complexity on ViT-H/14.}}
\label{tabel:Time_complexity}
\tiny
\setlength{\tabcolsep}{0.5mm}
\begin{tabular}{@{}ccc@{}}
\toprule
\multirow{1}{*}{\textbf{Method (↓)}}  &  \multirow{1}{*}{\textbf{w/o LwPTV}} & \multirow{1}{*}{\textbf{w/ LwPTV}} \\
\midrule

Task Arithmetic \cite{ilharco2022editing}
&  57.5 &  74.9   \\ 
Ties \cite{yadav2023resolving}
& 147.2s &  214.8s   \\  

PCB \cite{du2024parameter}
& 1861.0s& 1928.6   \\  
 
\bottomrule
\end{tabular}
\end{wraptable}

To investigate whether our method incurs significant computational overhead for layer-wise pruning on larger models, we conducted a time complexity analysis based on the ViT-H/14 architecture. Suppose the model has $L$ layers, the $l$-th layer has $d_l$ parameters, and $K$ models need to be merged:
(1) The time complexity of calculating $\mathbb{E} \left( \left| \boldsymbol{\tau }_{k}^{l}-\frac{1}{K}\sum_{k=1}^K{\boldsymbol{\tau }_{k}^{l}} \right| \right)$ is $\mathcal{O} (KLd_l)$.
(2) When the $l$-th layer of the final mask $\hat{m}$ is 0, we directly set the $l$-th layer parameters of the merged task vector to 0. In the worst case, all parameters need to be set to 0. In this case, the time complexity is $\mathcal{O} (Ld_l)$. So the total time complexity is $\mathcal{O} (KLd_l)$. 
In addition, we list the time overhead of Task Arithmetic, Ties and PCB with and without LwPTW on ViT-H/14 in Tab.\ref{tabel:Time_complexity}. The results show that whether we introduce ours into Task Arithmetic, Ties or PCB, it takes no more than 67.6 seconds. The time introduced by ours is acceptable or can even be ignored, especially for the model merging algorithms, which are relatively time-consuming such as PCB.

\subsection{Computational resources}
\label{Computational Resources and Runtimes}



Following Ilharco et al. \cite{ilharco2022editing}, we fine-tune separate instances of the pretrained ViT-H/14 on eight distinct datasets: Cars, DTD, EuroSAT, GTSRB, MNIST, RESISC45, SUN397, and SVHN. Training employs the AdamW optimizer with a learning rate of 0.001 and a batch size of 128. To ensure convergence while mitigating overfitting, dataset-specific training epochs are assigned: 35 epochs for Cars, 76 for DTD, 12 for EuroSAT, 11 for GTSRB, 5 for MNIST, 15 for RESISC45, 14 for SUN397, and 4 for SVHN.  All implementations use PyTorch on NVIDIA A800 GPUs.

\begin{table}[!htp]
\centering
\caption{Performance comparison for ID and OOD datasets on T5-large.}
\label{T5-large-OOD-T0-held-out}
\tiny
\setlength{\tabcolsep}{0.5mm}
\resizebox{\textwidth}{!}{%
\begin{tabular}{cccccccccccc}
\toprule
  \multirow{1}{*}{\textbf{Dataset (→)}}  &  \multicolumn{1}{c}{\textbf{In Domain}} & \multicolumn{9}{c}{\textbf{Out of Domain}} &\multirow{2}{*}{ \textbf{H-score}} \\ 
  \cmidrule(lr){2-2} \cmidrule(lr){3-11}  
 \multirow{1}{*}{\textbf{Method (↓)}} & \textbf{Avg.} & \textbf{rte} & \textbf{cb} & \textbf{wic} & \textbf{copa} & \textbf{h-swag}& \textbf{anli-r1}& \textbf{anli-r2}& \textbf{anli-r3} & \textbf{Avg.}  \\ 
\midrule
Pretrained &44.9 & 53.1 & 33.9& 51.4& 55.0& 30.5& 32.3& 35.4& 32.8 &40.6
 & 42.7 \\ 

Individual & 85.6 & 56.0  & 39.8 & 51.1 &58.3 & 29.6& 32.5  & 34.1 &33.2 & 41.8 &   56.2 \\

Weight Averaging \cite{wortsman2022model} &60.5  & 47.7& 48.2& 49.5& 55.0& 30.1& 33.7& 34.9& 33.3& 41.5& 49.2  \\ 
\midrule

Task Arithmetic \cite{ilharco2022editing}  & 73.3 &48.0& 50.0& 49.7& \textbf{73.0}& 28.6& \textbf{34.1}& 34.9& \textbf{33.1} & 43.9 & 54.9    \\ 

\rowcolor{gray!10} \textbf{w/ Ours}  & \textbf{76.3} &\textbf{66.1}& \textbf{57.1}& \textbf{50.3}& 71.0& \textbf{28.7} & 31.5& \textbf{34.9}& 33.0&\textbf{46.6}\textsuperscript{\best{(+2.7)}}&\textbf{57.9} \textsuperscript{\best{(+3.0)}}    \\ 

\midrule

Ties \cite{yadav2023resolving} &  71.0& 64.3& 62.5& \textbf{52.5}& 70.0& 30.2& \textbf{33.9}& 35.8& \textbf{35.0}& 48.0 & 57.3 \\

\rowcolor{gray!10} \textbf{w/ Ours}  &\textbf{71.9}& \textbf{67.9} & \textbf{67.9}& 52.4& \textbf{74.0}& \textbf{31.2}& 33.0& \textbf{37.2}& 34.8
& \textbf{49.8} \textsuperscript{\best{(+1.8)}} &  \textbf{58.8} \textsuperscript{\best{(+1.5)}} \\ 

\midrule

PCB-MERGING \cite{du2024parameter} &\textbf{75.9} &70.0& \textbf{64.3}& 51.7& 73.0& 30.6& \textbf{35.1}& \textbf{36.4}& \textbf{34.3} & 49.4 & 59.8    \\

\rowcolor{gray!10} \textbf{w/ Ours}  &75.8 &\textbf{70.0}& 62.5& \textbf{52.2}& \textbf{78.0}& \textbf{31.2}& 35.0& 36.3& 33.7 & \textbf{49.9} \textsuperscript{\best{(+0.5)}} & \textbf{60.2} \textsuperscript{\best{(+0.4)}}\\

\bottomrule
\end{tabular}
}
\end{table}

\begin{table}[H]
\centering
\caption{Performance comparison for ID  datasets on T5-large.}
\label{T5-large-T0}
\setlength{\tabcolsep}{0.8mm}
\resizebox{\textwidth}{!}{%
\begin{tabular}{ccccccccc}
\toprule
 \multirow{1}{*}{\textbf{Method (↓)}} & \textbf{PAWS} & \textbf{QASC} & \textbf{QuaRTz} & \textbf{StoryCloze} & \textbf{WikiQA} & \textbf{Winogrande}& \textbf{WSC}&  \textbf{Avg.} \\ 
\midrule

Pretrained  &48.1&33.8&53.1&47.0&37.9&50.9&43.3  &44.9  \\ 

 Individual &95.4 &97.6 &91.9 &90.4 &95.8 &79.2 &51.0 & 85.9 \\

\midrule
Weight Averaging \cite{wortsman2022model} &55.9&71.6 &54.4 &56.4&76.7 & 53.4&54.8 & 60.5\\ 
\midrule

Task Arithmetic \cite{ilharco2022editing} &  64.6& 74.4& 75.0& 80.8& \textbf{92.8}& 63.1& \textbf{62.5}&73.3 \\ 

\rowcolor{gray!10} \textbf{w/ Ours}  & \textbf{85.5} & \textbf{77.4} & \textbf{77.9} & \textbf{81.0} & 85.7& \textbf{67.2}& 59.6&\textbf{76.3} \textsuperscript{\best{(+3.0)}}   \\ 
\midrule

Ties \cite{yadav2023resolving} &92.6& 62.6& 75.0& 78.5& \textbf{58.7}& 75.7& 53.8& 71.0\\

  \rowcolor{gray!10} \textbf{w/ Ours}   &\textbf{92.8}& \textbf{67.0}& \textbf{77.1}& \textbf{78.8}& 55.6& \textbf{76.3} & \textbf{55.8}&\textbf{71.9} \textsuperscript{\best{(+0.9)}}\\

\midrule

PCB-MERGING \cite{du2024parameter} & \textbf{92.3} &78.4& \textbf{77.3}& \textbf{76.8}& \textbf{70.8}& \textbf{74.4}& 61.5& \textbf{75.9} \\ 

\rowcolor{gray!10} \textbf{w/ Ours}  &92.2& \textbf{79.0}& 76.6& 76.0& 69.9& 74.3& \textbf{62.5}&75.8 \textsuperscript{\color[RGB]{255,0,0}{(-0.1)}} \\ 
\bottomrule
\end{tabular}
}
\end{table}

\subsection{NLP task}

To evaluate the performance of our method on NLP tasks, we conducted experiments using T5-Large-LM-Adapt as the base model. We selected PAWS, QASC, QuaRTz, StoryCloze, WikiQA, Winogrande, and WSC as the ID datasets \cite{tammerging}, and rte, cb, wic, copa, h-swag, anli-r1, anli-r2, and anli-r3 as the OOD datasets \cite{sanh2022multitask}.
As shown in Tab.\ref{T5-large-OOD-T0-held-out}, in contrast to image classification tasks, the merged model demonstrates significantly better OOD performance than the pretrained model in the field of NLP. This superiority can be attributed to the high degree of semantic space overlap across different tasks, domains, and fields in NLP. This overlap implies that the  ID task vectors contain information that is beneficial for OOD tasks \cite{arora2021types, lewis2021question, elangovan2021memorization}. In other words, the discriminative patterns of OOD tasks have a non-zero mapping onto the discriminative patterns of ID tasks \cite{LZZC25}. Under these circumstances, enhancing the OOD performance of the merged model requires amplifying the role of task-specific parameters within the ID task, which can be achieved by pruning redundant parameters.
Unlike images, which are spatial and have higher redundancy with many local blocks containing overlapping information, allowing for robust compression by deleting entire Transformer layers while preserving important global information and reducing overfitting to high-frequency noise or specific dataset artifacts \cite{chen2022principle, hou2022multi}, language is sequential with meanings that are compositional and context-sensitive. Discarding entire layers might disrupt the sentence-level or token-level dependencies crucial for semantic understanding \cite{sajjad2023effect, ding2025revisiting}. Therefore, for NLP tasks, we perform parameter pruning at the parameter level.
As shown in Tab.\ref{T5-large-OOD-T0-held-out}, compared to the baseline methods Task Arithmetic, Ties, and PCB, our method achieves relative improvements of 3\%, 1.5\%, and 0.4\% on the H-score, respectively. The detailed results on the ID dataset are presented in Tab.\ref{T5-large-T0}.
Following Ilharco et al. \cite{ilharco2022editing, yadav2023resolving, du2024parameter, tammerging}, for Task Arithmetic, we set the range of $\lambda$ to $[0.1, 1.5]$ with a step size of 0.1; for Ties and PCB, we searched over mask ratios $r \in \{0.05, 0.1, 0.2\}$ and set the range of $\lambda$ to $[0.1, 2.5]$ with a step size of 0.1. For our method, we set the range of the pruning ratio $\eta$ to $[0.1, 0.7]$ with a step size of 0.1.

\section{Additional visualization results }
\label{Visualization Results}

\subsection{T-SNE visualization}  
\label{sec:T-SNE}

\begin{figure}[ht]
  \centering

  \begin{subfigure}[b]{0.48\textwidth}
    \includegraphics[width=\linewidth]{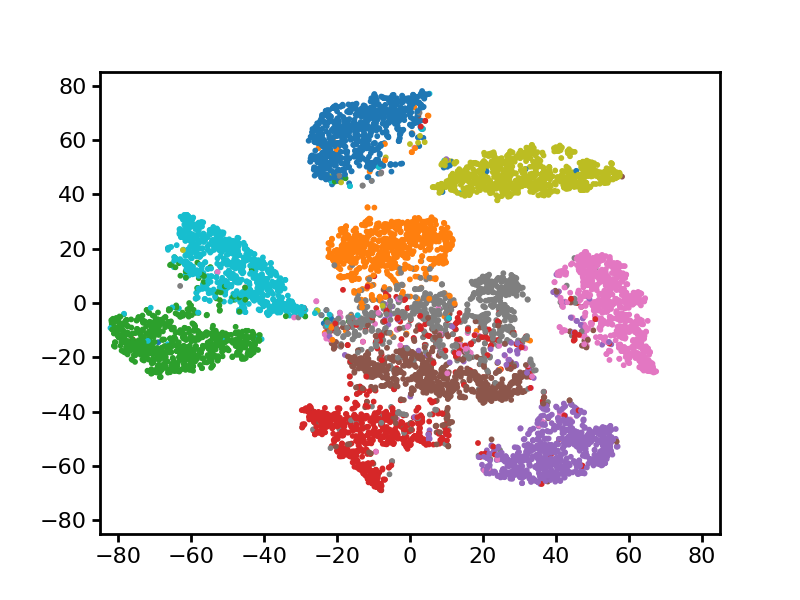}
    \caption{STL10: Task Arithmetic}
    \label{fig:stl10_no}
  \end{subfigure}
  \hfill
  \begin{subfigure}[b]{0.48\textwidth}
    \includegraphics[width=\linewidth]{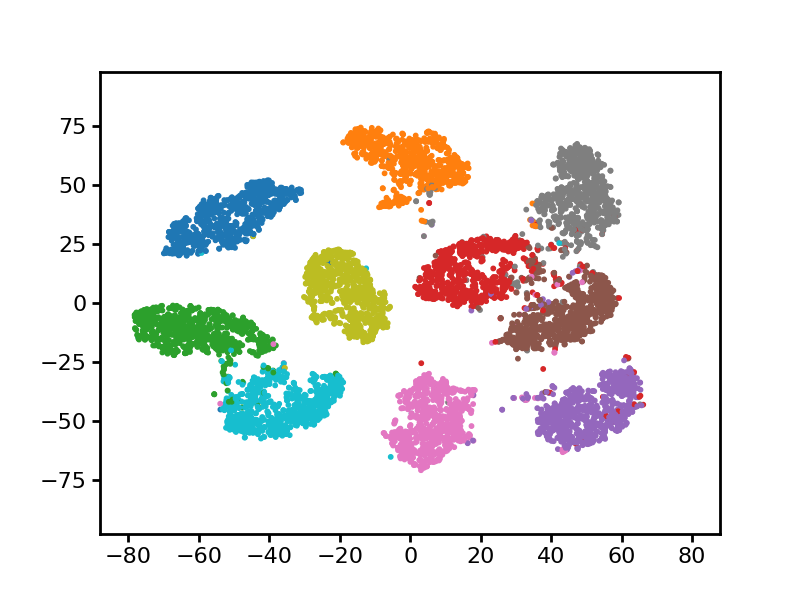}
    \caption{STL10: Task Arithmetic w/ \method}
    \label{fig:stl10_method}
  \end{subfigure}

  \caption{T-SNE visualizations on STL10 comparing Task Arithmetic without and with \method.}
  \label{fig:tsne_all}
\end{figure}

To explore the effectiveness of ours intuitively, we also present a T-SNE visualization for Task Arithmetic and the Task Arithmetic+\method on OOD tasks, where STL10 datasets are shown in \ref{fig:tsne_all}. 
Compared to Task Arithmetic, our  \method demonstrates significantly enhanced the separability of representations across different categories. Furthermore, the intra-class clustering exhibits greater compactness when employing our approach. It indicates that introducing our proposed mask vector to prune the task vectors \method can extract discriminative features and thus enhances the generalization ability of the merged model.

\subsection{Saliency score and mask vectors}
\label{sec:SSMV}

\begin{figure}[H]
    \centering
    \vspace{-1.7em}
    \includegraphics[width=0.9\textwidth]{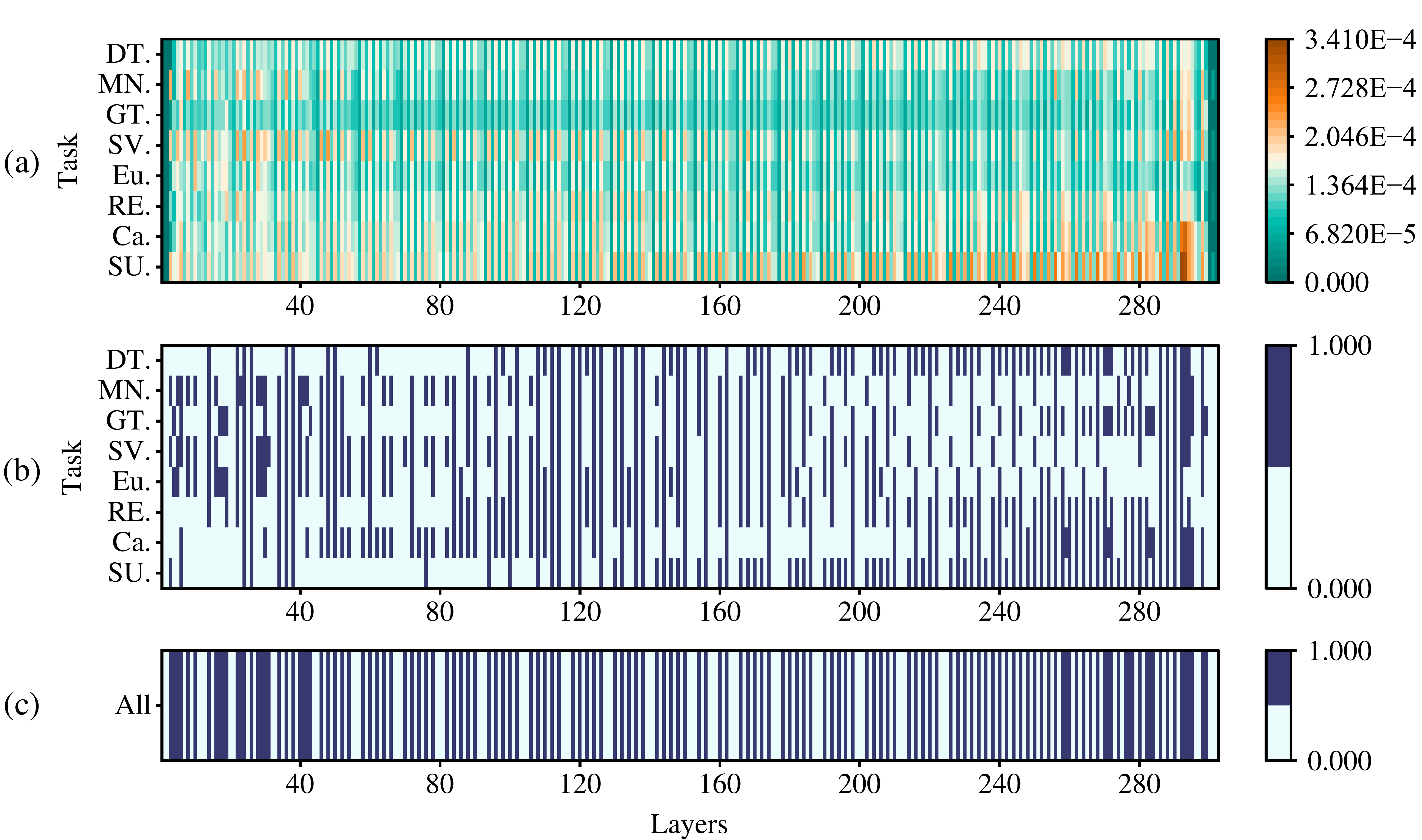}  
    \vspace{-5pt}
    \caption{\small{Visualizing of (a) salience score matrix, (b) mask vector for each task, and (c) final mask vector, all on VIT-L/14, where x-axis denotes the layer index, y-axis in (a-b) denotes task name.}}
        \vspace{-10pt}
    \label{fig:Significance score ViT-L-14}
\end{figure}

\begin{figure}[H]
    \centering
    \vspace{-1.7em}
    \includegraphics[width=0.9\textwidth]{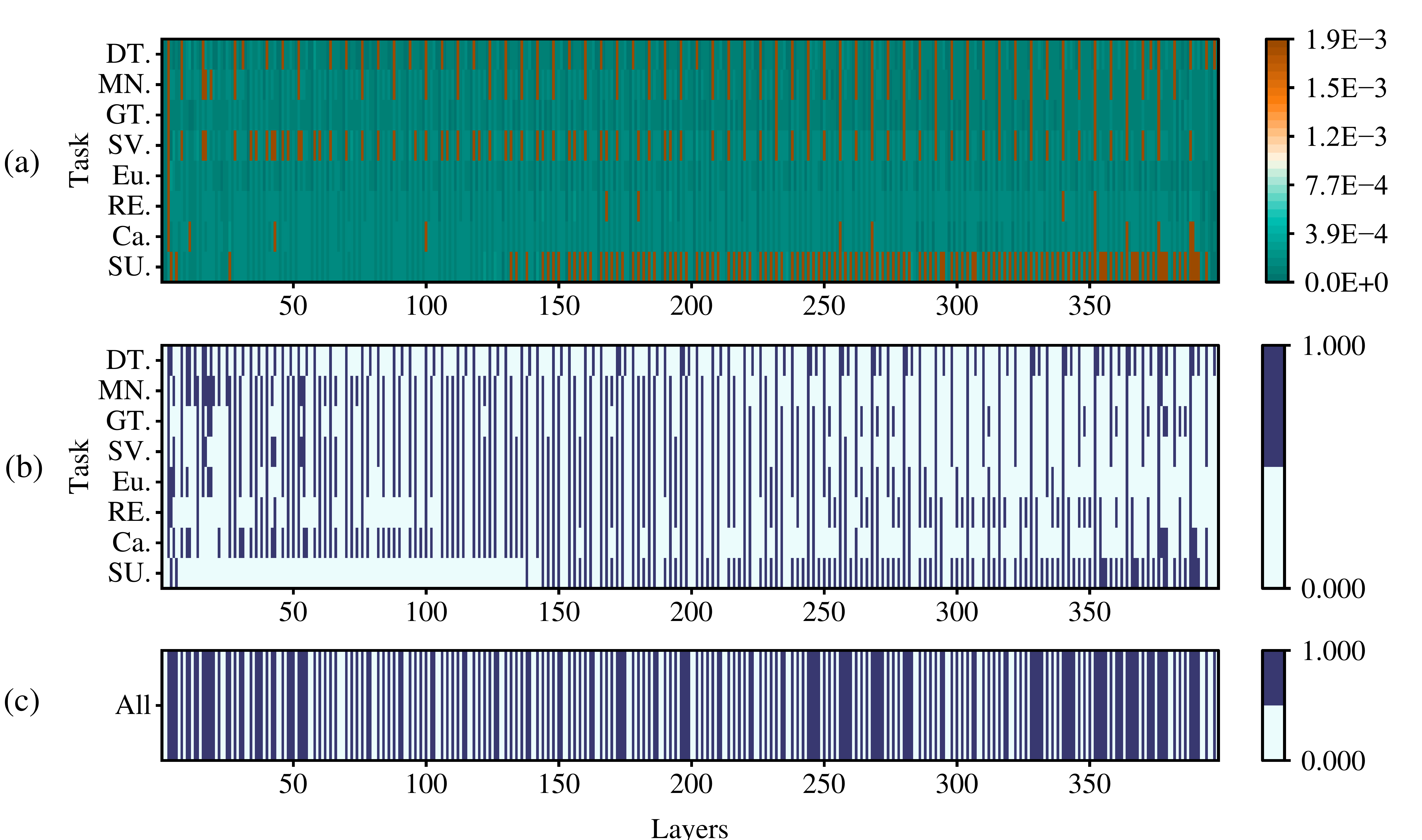}  
    \vspace{-5pt}
    \caption{\small{Visualizing of (a) salience score matrix, (b) mask vector for each task, and (c) final mask vector, all on VIT-H/14, where x-axis denotes the layer index, y-axis in (a-b) denotes task name.}}
        \vspace{-10pt}
    \label{fig:Significance score ViT-H-14}
\end{figure}

To facilitate an intuitive analysis of saliency score, masks associated with each task vector and their interrelationships, we also visualize them for ViT-L/14 in Fig.\ref{fig:Significance score ViT-L-14} and ViT-H/14 in Fig.\ref{fig:Significance score ViT-H-14}. 

\section{Discussion}
\label{Discussion}

\subsection{Limitation and future work}
\label{sec:limi_and_fut}
Although our method offers valuable insights into the OOD performance of merged models, certain limitations warrant consideration. Firstly, akin to prior approaches, \method is contingent upon identical model architectures and shared initializations, thereby restricting its broad applicability across diverse model paradigms. Secondly, it may compromise ID performance when employed with certain model merging techniques. Consequently, investigating model merging methodologies that transcend architectural disparities, circumventing the current reliance on homogeneous architectures, and devising novel pruning strategies that enhance OOD performance while preserving or augmenting ID performance, constitute promising avenues for future research.

\subsection{Broader impacts}
\label{sec:impact}
Our method enhances the OOD performance of the merged model without requiring access to expert data. This contributes to the preservation of privacy for expert data and reinforces data security, thereby exerting a positive impact on societal data security. Our method does not exhibit any significant negative societal impacts.

\newpage
\section*{NeurIPS Paper Checklist}

\begin{enumerate}

\item {\bf Claims}
    \item[] Question: Do the main claims made in the abstract and introduction accurately reflect the paper's contributions and scope?
    \item[] Answer: \answerYes{}
    \item[] Justification: As shown in Section \ref{sec:Introduction}. 

\item {\bf Limitations}
    \item[] Question: Does the paper discuss the limitations of the work performed by the authors?
    \item[] Answer: \answerYes{}
    \item[] Justification: The limitations of the work are shown in Section \ref{sec:limi_and_fut}

\item {\bf Theory assumptions and proofs}
    \item[] Question: For each theoretical result, does the paper provide the full set of assumptions and a complete (and correct) proof?
    \item[] Answer:  \answerYes{}
    \item[] Justification: As shown in Section \ref{sec:observation} and Appendix \ref{sec:Proposition 1}.

    \item {\bf Experimental result reproducibility}
    \item[] Question: Does the paper fully disclose all the information needed to reproduce the main experimental results of the paper to the extent that it affects the main claims and/or conclusions of the paper (regardless of whether the code and data are provided or not)?
    \item[] Answer: \answerYes 
    \item[] Justification: We provide the dataset details in Appendix \ref{sub:datasets} and implementation details in Section \ref{sec:Experiments} and Appendix \ref{Computational Resources and Runtimes} to reproduce the main experimental results. 

\item {\bf Open access to data and code}
    \item[] Question: Does the paper provide open access to the data and code, with sufficient instructions to faithfully reproduce the main experimental results, as described in supplemental material?
    \item[] Answer: \answerYes 
    \item[] Justification: We have released the code and experiment setting details in our supplemental material.

\item {\bf Experimental setting/details}
    \item[] Question: Does the paper specify all the training and test details (e.g., data splits, hyperparameters, how they were chosen, type of optimizer, etc.) necessary to understand the results?
    \item[] Answer: \answerYes{} 
    \item[] Justification: We provide the dataset details in Appendix \ref{sub:datasets}, implementation details in Appendix \ref{Computational Resources and Runtimes} and hyperparameter details in Section \ref{sec:Experiments}.

\item {\bf Experiment statistical significance}
    \item[] Question: Does the paper report error bars suitably and correctly defined or other appropriate information about the statistical significance of the experiments?
    \item[] Answer: \answerYes{} 
    \item[] Justification: Due to the inherent variability introduced by different random seeds in random mask and DARE, we report the optimal results obtained using three distinct random seeds during the mask generation process in Appendix \ref{sec:Pruning}. 

\item {\bf Experiments compute resources}
    \item[] Question: For each experiment, does the paper provide sufficient information on the computer resources (type of compute workers, memory, time of execution) needed to reproduce the experiments?
    \item[] Answer: \answerYes{} 
    \item[] Justification: As shown in Appendix \ref{Computational Resources and Runtimes}.
    
\item {\bf Code of ethics}
    \item[] Question: Does the research conducted in the paper conform, in every respect, with the NeurIPS Code of Ethics \url{https://neurips.cc/public/EthicsGuidelines}?
    \item[] Answer: \answerYes{} 
    \item[] Justification: This research is conducted in the paper conform, with the NeurIPS Code of Ethics.

\item {\bf Broader impacts}
    \item[] Question: Does the paper discuss both potential positive societal impacts and negative societal impacts of the work performed?
    \item[] Answer: \answerYes 
    \item[] Justification: As shown in Appendix \ref{sec:impact}.
    
\item {\bf Safeguards}
    \item[] Question: Does the paper describe safeguards that have been put in place for responsible release of data or models that have a high risk for misuse (e.g., pretrained language models, image generators, or scraped datasets)?
    \item[] Answer: \answerNA{} 
    \item[] Justification:

\item {\bf Licenses for existing assets}
    \item[] Question: Are the creators or original owners of assets (e.g., code, data, models), used in the paper, properly credited and are the license and terms of use explicitly mentioned and properly respected?
    \item[] Answer: \answerYes{} 
    \item[] Justification: We adequately cite all of the used datasets and architectures.

\item {\bf New assets}
    \item[] Question: Are new assets introduced in the paper well documented and is the documentation provided alongside the assets?
    \item[] Answer: \answerNA{} 
    \item[] Justification: While we provide a codebase that is extensible and modular enough to be reused by many researchers in the field, we are not releasing any libraries or datasets.

\item {\bf Crowdsourcing and research with human subjects}
    \item[] Question: For crowdsourcing experiments and research with human subjects, does the paper include the full text of instructions given to participants and screenshots, if applicable, as well as details about compensation (if any)? 
    \item[] Answer: \answerNA{} 
    \item[] Justification: 

\item {\bf Institutional review board (IRB) approvals or equivalent for research with human subjects}
    \item[] Question: Does the paper describe potential risks incurred by study participants, whether such risks were disclosed to the subjects, and whether Institutional Review Board (IRB) approvals (or an equivalent approval/review based on the requirements of your country or institution) were obtained?
    \item[] Answer: \answerNA{} 
    \item[] Justification: 

\item {\bf Declaration of LLM usage}
    \item[] Question: Does the paper describe the usage of LLMs if it is an important, original, or non-standard component of the core methods in this research? Note that if the LLM is used only for writing, editing, or formatting purposes and does not impact the core methodology, scientific rigorousness, or originality of the research, declaration is not required.
    \item[] Answer: \answerNA{} 
    \item[] Justification: 

\end{enumerate}

\end{document}